\pdfoutput=1
\documentclass[11pt]{article}
\usepackage[T1]{fontenc}
\usepackage{mathpazo} % or any other package that provides a cursive font
\usepackage{amsmath}
\usepackage[hyphens]{url} % This line is now effectively taking care of URL options
\usepackage{hyperref}
\hypersetup{
    colorlinks=true,
    linkcolor=blue,
    filecolor=magenta,      
    urlcolor=cyan,
    breaklinks=true
}

\usepackage[english]{babel}
\usepackage{hyphenat}

\usepackage[]{latex/acl}

\usepackage{times}
\usepackage{latexsym}
\usepackage{graphicx}
\usepackage{ulem}
\usepackage[utf8]{inputenc}

\usepackage{caption}
\usepackage{subcaption}

\usepackage{longtable}
\usepackage{multirow}
\usepackage{supertabular}

\usepackage{microtype}

\usepackage{inconsolata}

\title{BEnQA: A Question Answering and Reasoning Benchmark for Bengali and English}

\author{
    Sheikh Shafayat, 
    H M Quamran Hasan, \\
    \textbf{Minhajur Rahman Chowdhury Mahim, 
    Rifki Afina Putri, 
    James Thorne,
    Alice Oh}
  \\
  \vspace{2 cm}
\begin{tabular}{c}
  KAIST, Republic of Korea\\
  \texttt{\{\href{mailto:sheikh.shafayat@kaist.ac.kr}{\color{black}{sheikh.shafayat}}, \href{mailto:quamranhasan@kaist.ac.kr}{\color{black}{quamranhasan}}, \href{mailto:minhaj@kaist.ac.kr}{\color{black}{minhaj}}, \href{mailto:rifkiaputri@kaist.ac.kr}{\color{black}{rifkiaputri}}, \href{mailto:thorne@kaist.ac.kr}{\color{black}{thorne}}\}@kaist.ac.kr}, \\
  \texttt{alice.oh@kaist.edu}
\end{tabular}
}

\begin{document}
\maketitle
\begin{abstract}
In this study, we introduce BEnQA\footnote{We release the dataset at: \url{https://github.com/sheikhshafayat/BEnQA}}, a dataset comprising parallel Bengali and English exam questions for middle and high school levels in Bangladesh. Our dataset consists of approximately 5K questions covering several subjects in science with different types of questions, including factual, application, and reasoning-based questions. We benchmark several Large Language Models (LLMs) with our parallel dataset and observe a notable performance disparity between the models in Bengali and English. We also investigate some prompting methods, and find that Chain-of-Thought prompting is beneficial mostly on reasoning questions, but not so much on factual ones. We also find that appending English translation helps to answer questions in Bengali. Our findings point to promising future research directions for improving the performance of LLMs in Bengali and more generally in low-resource languages.
\end{abstract}

\section{Introduction} % Sheikh
Large language models (LLMs) like GPT-4 have shown impressive performance in many complex natural language processing tasks including reasoning and question answering, all of which have been subject to intense research in recent years \cite{bang2023multitask, liu2023evaluating}. However, most of this research has focused on English and other high-resource languages, often neglecting medium- to low-resource languages. To make matters more difficult, most of the benchmarks designed for measuring progress in LLM research are only available in English and a handful of other languages, making it very hard to compare different LLM skills in low-resource languages.

This issue is particularly critical in light of the growing use of LLMs like ChatGPT in the context of education \cite{khanacademy2023}. The absence of LLMs that perform equally well in non-English languages would lead to further divide in access to education and technology, highlighting a critical need for more inclusive language model development.

Addressing this gap, our work specifically targets Bengali, a language spoken by 272 million speakers worldwide \cite{zeidan_languages_2023}, yet considered a low-resource language. We make the following contributions:
\begin{itemize}
    \item We introduce BEnQA, a science question-answering dataset in English and Bengali, consisting of 5,161 questions taken from the Bangladeshi national curriculum for grade school exams. This dataset covers a wide range of subjects and question types, including those requiring multi-step reasoning.
    \item Our dataset is parallel in English and Bengali, allowing us to benchmark existing LLMs and measure the performance gap between them. Additionally, the parallel nature of our dataset ensures a fairer comparison between the two languages.
    \item Through our evaluation of various LLMs, we indeed observe a significant performance discrepancy between questions in English and Bengali. Additionally, we investigate the impact of different prompting techniques on LLMs' performance in Bengali, such as Chain-of-Thought (CoT) prompting and appending English translation to the prompt.
\end{itemize}

\section{Related Work} % Sheikh, Minhaj

\subsection*{Multilingual Reasoning Benchmark}
There have been many English benchmarks to evaluate the reasoning capabilities of LLMs, such as COPA \cite{roemmele2011choice}, HellaSwag \cite{zellers-etal-2019-hellaswag}, CosmosQA \cite{huang-etal-2019-cosmos}, and CommonsenseQA \cite{talmor-etal-2019-commonsenseqa}. For non-English languages, one of the widely used benchmarks is X-COPA \cite{ponti-etal-2020-xcopa}. A recent work \cite{doddapaneni-etal-2023-towards} provides a human translation of this dataset in several Indic languages, including Bengali. 
Besides these, BIG-Bench Hard \cite{suzgun2022challenging} is a dataset consisting of a collection of 23 challenging BIG-Bench tasks \cite{srivastava2022beyond} that measure a model's reasoning ability across various tasks such as logical deduction, multi-step arithmetic, and more.

Another line of work develops datasets using academic exam questions. Notable examples include MMLU \cite{hendrycks2020measuring}, which covers multi-task problems from grade school to college level; MATH \cite{hendrycks2021measuring}, consisting of competitive math problems; GSM8k \cite{cobbe2021training}, which consists of grade school mathematics problems; and ARC dataset \cite{clark2018think} that proposes grade school level multiple-choice science questions. Recent GPT-4 technical report \cite{openai2023gpt4} also included benchmark results of many exams such as GRE, bar exam, and AP exam. So did some other works such as \citet{kung2023performance} and \citet{choi2023chatgpt}. 

However, most of these academic datasets are focused on English, with limited availability in other languages. One such example is MGSM \cite{shi2022language}, a professionally translated grade school math problem from GSM8k in 10 languages, one of which is Bengali. For Bengali, this is, in fact, the only commonly used academic reasoning dataset. One example of a non-English academic exam benchmark dataset is IndoMMLU \cite{koto2023large}, which consists of questions from various exam subjects collected by professional teachers.  Our work is very similar to this in spirit.

\subsection*{Multilingual Prompting}
Chain-of-Thought (CoT) prompting \cite{wei2022emergent,wei2023chainofthought} has ushered a new wave of work in eliciting reasoning behavior in large language models. \citet{shi2022language} demonstrated the effectiveness of CoT and reported an increase in performance in mathematical reasoning while doing step-by-step reasoning in non-English languages. The same work also reports that translating the questions using Google Translate and then doing step-by-step reasoning in English often works even better. \citet{ahuja2023mega} also used a similar approach. 
More recently, \citet{huang2023not} introduced cross-lingual thought prompting that first translates the original query into English and then does CoT reasoning in English. 

However, one obvious issue with doing CoT reasoning in English is that it renders the LLM output less useful for non-English users. Especially, when LLMs are used in educational contexts, CoT often provides helpful hints for learners \cite{han2023recipe}. In our work, we demonstrate that it is possible to do CoT reasoning in the target language and improve performance, as long as we have access to the English translation of the questions. 

\begin{figure}[t]
    \centering
    \resizebox{\columnwidth}{!}{\includegraphics{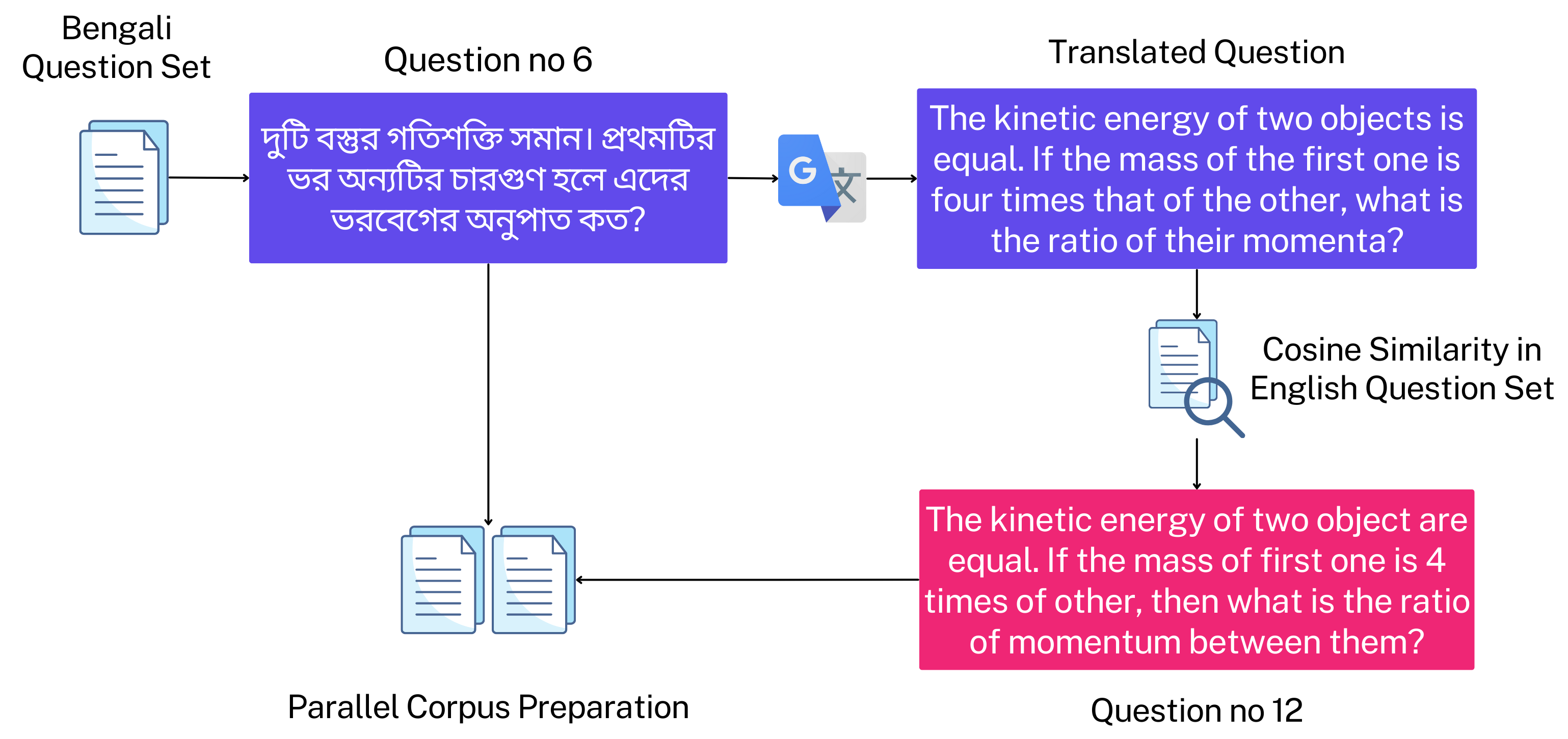}}
    \caption{Our pipeline for creating parallel corpus in BEnQA.}
    \label{fig:corpus_making}
\end{figure}

\begin{figure*}[t]
    \centering
    \begin{subfigure}[b]{0.4\textwidth}
        \centering
        \includegraphics[width=\textwidth]{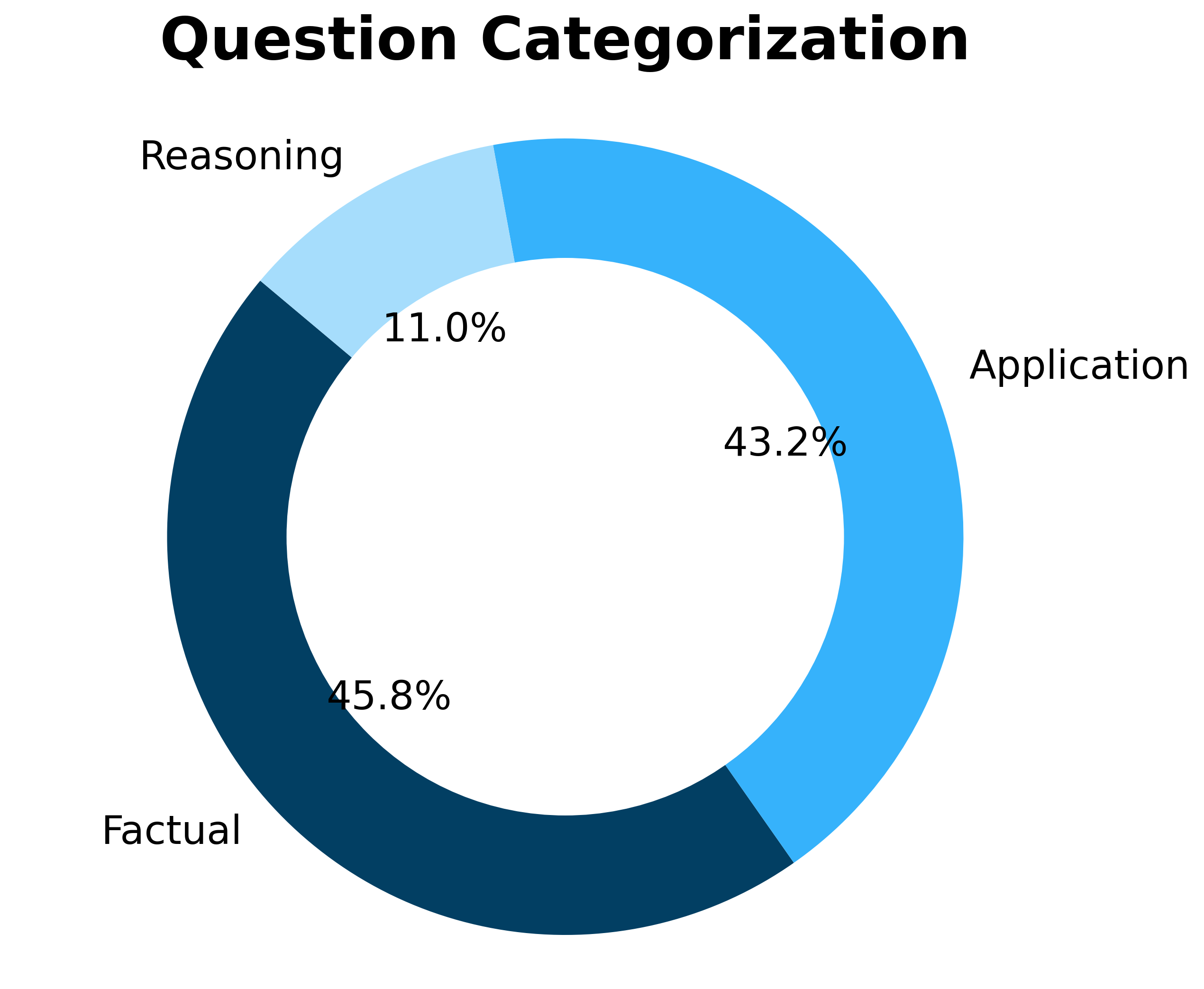}
        \caption{Statistics by categories.}
        \label{fig:que_cat}
    \end{subfigure}
    \hfill
    \begin{subfigure}[b]{0.5\textwidth}
        \centering
        \includegraphics[width=\textwidth]{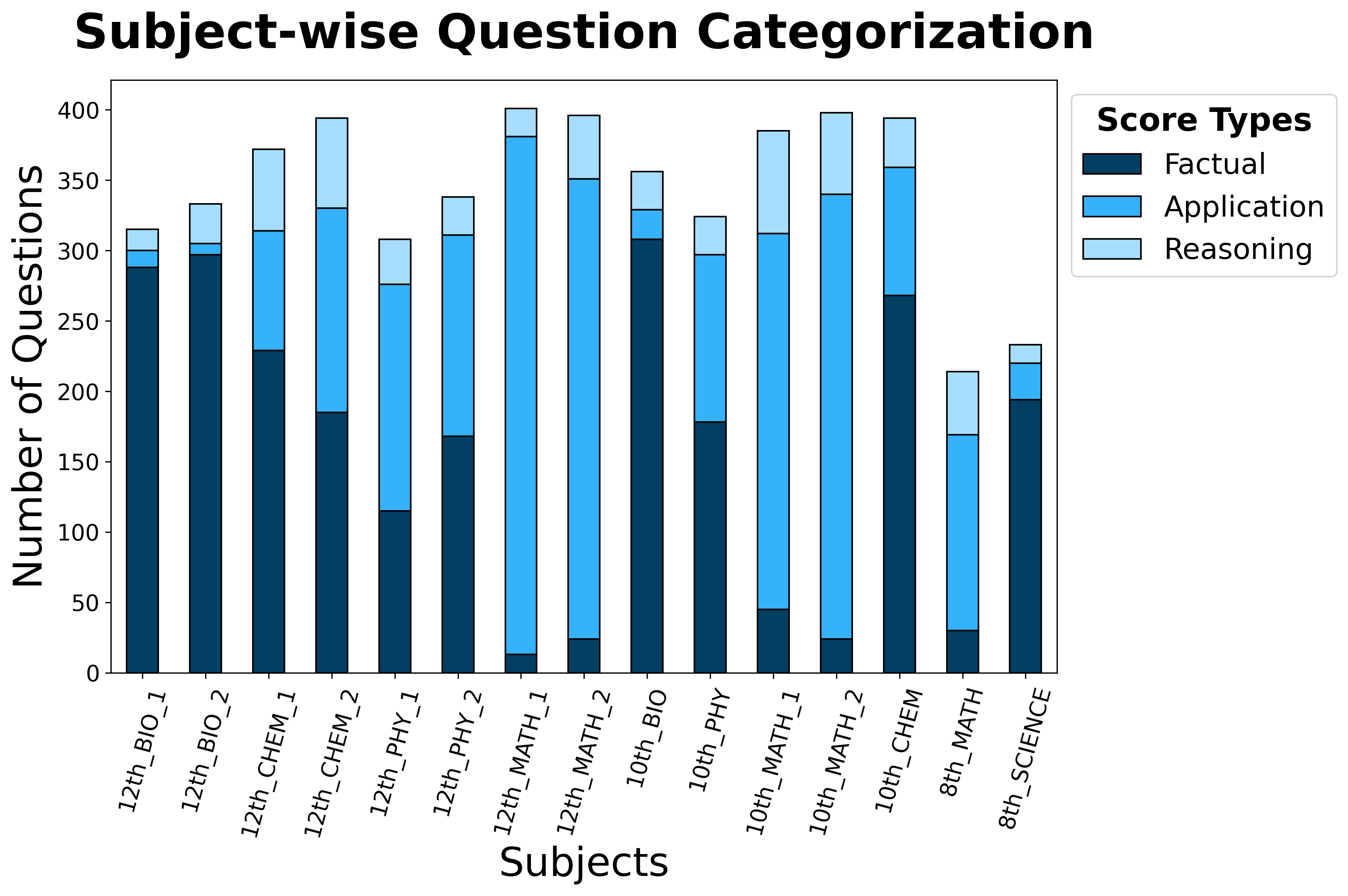}
        \caption{Statistics by grade and subject.}
        \label{fig:sub_cat}
    \end{subfigure}
    \caption{Statistics of our BenQA dataset.}
\end{figure*}

\section{BEnQA Dataset} %Minhaj

Our dataset, BEnQA, is a parallel corpus of English-Bengali science questions in a multiple-choice format from 8th, 10th, and 12th grade exams. These questions are sourced from nationwide board exams in Bangladesh and are officially available in both English and Bengali, giving us a high-quality parallel corpus. 

\subsection{Dataset Curation} % Minhaj
Examination questions in Bangladesh are predominantly available in print rather than digital format. Therefore, we collect examination papers and use readily accessible solution books to create the ground truth for the questions. We then employ four typists, proficient in both Bengali and English, to digitize the questions and their corresponding answers. This process enables the conversion of questions and their answers into a digital format with minimal errors.

All questions are of the multiple-choice format, each presenting four options. The typists are instructed to format mathematical equations and chemical formulas using \LaTeX{}. Questions involving figures are excluded. Additionally, we apply specific heuristics to filter out noisy data, annotation errors, and poor-quality questions, including those with inconsistencies between the ground truths in English and Bengali.

\subsection{Parallel Corpus Creation}
Typically, questions in Bangladeshi national board exams are initially formulated in Bengali and then translated into English. This process allows us to create a parallel dataset. However, the question order in each version does not always align (i.e., the first question in the Bengali version can be the eleventh question in the English version). To address this, we develop a very simple yet effective algorithm for aligning the English and Bengali questions, illustrated in Figure \ref{fig:corpus_making}. 

First, we translate each Bengali question to English via Google Translate API. Within each subject, we then match the Google-translated version with the actual English translation of the question, using the cosine similarity of their embeddings. We use OpenAI \texttt{text-embedding-ada-002} embedding to achieve this. After the initial matching, two native Bengali speakers who are also proficient in English, manually verify each question to ensure that each Bengali question corresponds to its English counterpart. We also filter out a few questions for which the ground truth is different in English and Bengali. Manual inspection reveals that such cases typically result from annotator errors or issues with the questions themselves.

Since these translations are often carried out by school teachers with varying levels of English proficiency, some English questions may contain subtle grammatical errors or sound unnatural to native English speakers. We investigate the impact of these grammatical inconsistencies on model performance by creating a grammar-corrected version of the English questions using GPT-4 with human supervision in the loop. Our findings indicate that grammatical errors generally do not significantly affect model performance (see Appendix \ref{appendix:grammar_mistakes}).

\subsection{Dataset Properties} % Minhaj
\label{sec:data_prop}
In our proposed BEnQA corpus, there is a total of 5,161 questions available in both English and Bengali. This dataset comprises 55\% (2,857) questions from the 12th grade, 36\% (1,857) from the 10th grade, and 9\% (447) from the 8th grade. The 12th-grade section encompasses various subjects like Mathematics, Physics, Chemistry, and Biology, further divided into part I and part II based on sub-topics. For the 10th grade, it includes Mathematics I, Mathematics II, Physics, Chemistry, and Biology. In the case of the 8th grade, the dataset includes Mathematics and Science, with the latter being a comprehensive subject covering 8th-grade science. Figure \ref{fig:sub_cat} summarizes the subject and grade-wise question statistics.

\subsection{Dataset Categorization}
Our dataset encompasses various question types, each demanding specific skills, irrespective of the subject or grade. Based on the skills necessary for solving them, we categorize the questions into three distinct types:

\begin{itemize}
    \item \textbf{Factual Knowledge:} These are questions that solely rely on knowledge of basic facts, events, concepts, dates, etc. Answering such questions does not involve any form of analysis or reasoning.
    \item \textbf{Procedural \& Application:} This category comprises questions that require the ability to apply a procedure, utilize a familiar concept, or employ a formula for solving.
    \item \textbf{Reasoning:} Questions falling into this category demand multiple steps of analysis or reasoning to arrive at a solution.
\end{itemize}

We utilize GPT-4 for the purpose of categorizing the questions into these groups. We use the prompt given in Table \ref{tab:categorizer} to categorize them. The categorization results are depicted in Figure \ref{fig:que_cat} for question-type distribution and  Figure \ref{fig:sub_cat} for subject-wise breakdown. Generally, the majority of questions in Biology and Chemistry lean towards factual knowledge. Conversely, Mathematics questions are primarily centered around procedural \& application skills, with some instances demanding reasoning. In the case of Physics, the number of factual and procedural questions is nearly equivalent.

% \input{prompting}
% \section{Translation Appended Prompting} 
% In this work, our objective is to capitalize on the proficiency of Language Models (LLMs) in English while ensuring that their responses are generated in the target language, which, in our case, is Bengali. To achieve this, we append the English translation to the input and instruct the models to exclusively utilize the English translation for context understanding. The models are prompted to consistently produce responses in the target language, as illustrated in Figure \ref{fig:prompt}. 

% We explored three sources for obtaining the translations: Google Translate, GPT-3.5 and GPT-4. Previous work \cite{jiao2023chatgpt} has shown that GPT-4 outperforms both GPT-3.5 and commercially available translation services: a manual inspection by native Bengali speakers also verified this claim. 
% \label{sec:exp_prompt}

\section{Experiment Setup} % Sheikh
To assess the performance of existing LLMs on our newly created dataset, we evaluate them using the BEnQA dataset across several open-source and proprietary models. Following the recommendation from prior work on ChatGPT \cite{lai2023chatgpt}, we keep the prompt in English for both Bengali and English datasets. 
Most of the benchmark results presented in the main paper are conducted in a zero-shot setting to save tokenization costs in proprietary models \cite{petrov2023language}. Additionally, experiments in 3-shot and 5-shot settings were performed as supplementary analyses, details of which are provided in Appendix \ref{appendix:additional results}.

\subsection{Models}  % Sheikh
To evaluate the current LLMs capability in our dataset, we use the following models:
\subsubsection*{English}
\begin{itemize}
    \item \textbf{Proprietary LLMs}: \texttt{gpt-4-1106-preview}, \texttt{gpt-3.5-turbo-1106}, \texttt{Claude 2.1}
    \item \textbf{Open-source LLMs:} \texttt{LLama-2 (7B)} \texttt{LLama-2 (13B)} and \texttt{Mistral (7B)}
\end{itemize}

\subsubsection*{Bengali}
\begin{itemize}
    \item We utilize the proprietary models mentioned above for our Bengali experiments. 
\end{itemize}

It is worth noting that most open-source models do not perform well on Bengali. For this reason, we conduct all open-source model experiments on only the English version of the dataset.

Most of the other experiments (i.e., ablations, exploration of prompting techniques, and others) in this paper are conducted using GPT-3.5 Turbo in order to avoid the high cost associated with more advanced proprietary models, particularly for Bengali \cite{ahia-etal-2023-languages,petrov2023language}. Opting for benchmarking on GPT-3.5 ensures a balanced trade-off between value and cost.

\subsection{Prompts}
We explore some prompt variations in our experiments, including prompts with Chain-of-Thought (CoT) and without CoT. Detailed descriptions of the specific prompts used in our experiments are provided in Appendix \ref{appendix: prompt}. Throughout all experiments, the model was explicitly instructed to do CoT reasoning in Bengali and refrain from using English in order to stick to our original goal of making the model output useful to Bengali users.

% We also explore four possible variations of this prompt method:
% \begin{itemize}
%     \item \textbf{Baseline:} Questions asked only in Bengali. 
%     \item \textbf{Gold Translation Append:} Gold English translation taken from the English question is appended to the Bengali question.
%     \item \textbf{GPT-3.5 Translation:} Same as before, but the English translation is done by GPT-3.5 model.
%     \item \textbf{GPT-4 Translation:} Same as before, but the English translation is done by GPT-4.
% \end{itemize}
% The premise is that since GPT-4 translation is of better quality than that of GPT-3.5 \cite{jiao2023chatgpt}, and probably not as good as human gold translation, it might give us a clear clue of the effect of translation quality on performance. 

\subsection{Evaluation Metric}
The evaluation process involves a manual assessment of model outputs to determine if the final answer aligns with the ground truth. It is important to note that our evaluation does not scrutinize the validity of intermediate reasoning steps. In other words, we consider the answer correct if the final result is accurate, without delving into the examination of the validity of each intermediate reasoning step.
% The few-shot experiment results in the appendix have been evaluated using automated evaluation. 

% \subsection{Extension to Other Datasets}
% \label{subsec:cot_lang_test}
% Additionally, to verify whether the effectiveness of this prompting strategy generalizes to other datasets, we perform the same experiment on Bengali COPA and BIG-Bench Hard datasets. For these datasets, our experiments are conducted following these settings:
% \begin{itemize}
%     \item \textbf{EN:} Asking only in English (using the original data). 
%     \item \textbf{BN:} Asking only in Bengali. As mentioned earlier, this translation is obtained through GPT-4, along with human supervision of the generated translations.
%     \item \textbf{BN + EN (Gold):} Asking in Bengali and appending the original data. 
%     \item \textbf{BN + EN (GPT-3.5):} Asking in Bengali and appending the English translation done by GPT-3.5.
%     \item \textbf{BN + EN (GPT-4):} Asking in Bengali and appending the English translation done by GPT-4.
% \end{itemize}
% All the experiments involving prompting with appended English translation are conducted exclusively on GPT-3.5(\texttt{gpt-3.5-turbo-1106}) due to the high cost associated with more advanced proprietary models, particularly for Bengali \cite{ahia-etal-2023-languages}. Opting for benchmarking on GPT-3.5 ensures a balanced trade-off between value and cost.

\begin{figure*}[t]
    \centering
    \begin{subfigure}[b]{0.49\textwidth}
        \centering
        \includegraphics[width=\textwidth]{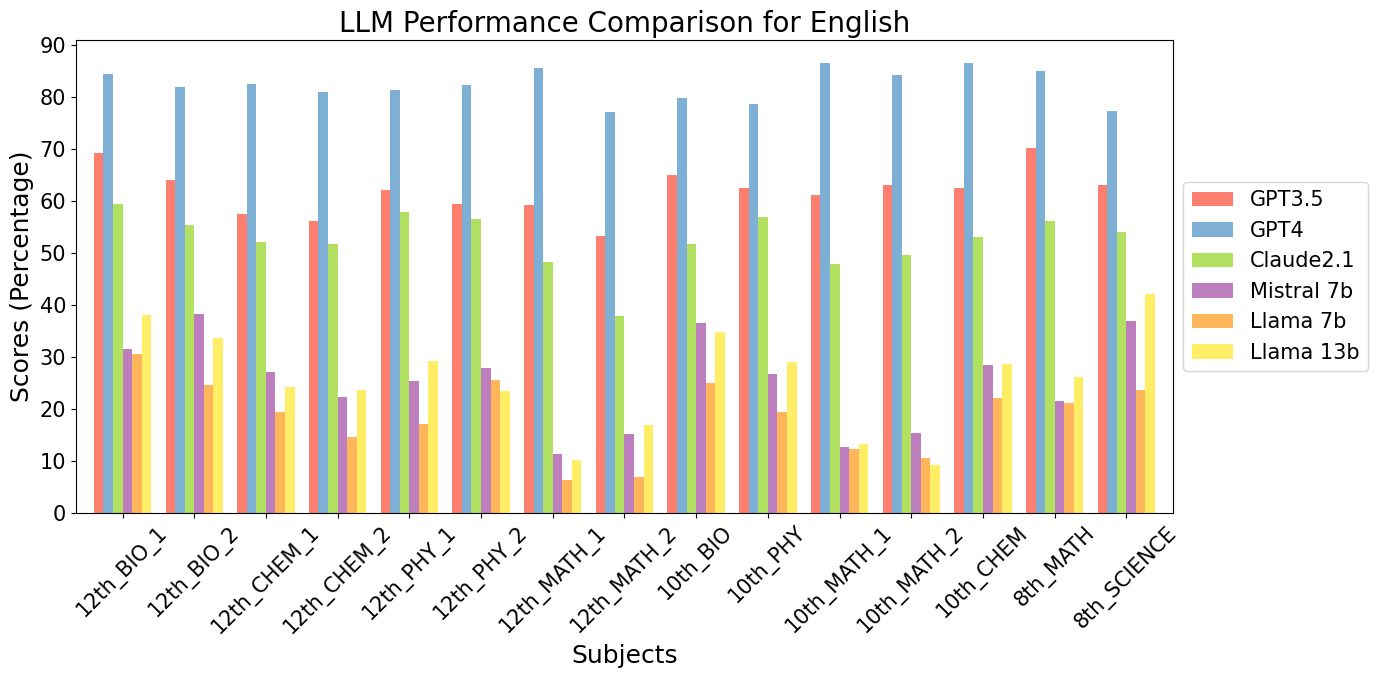}
        \caption{LLMs results in English questions}
        \label{fig:all_en}
    \end{subfigure}
    \hfill
    \begin{subfigure}[b]{0.49\textwidth}
        \centering
        \includegraphics[width=\textwidth]{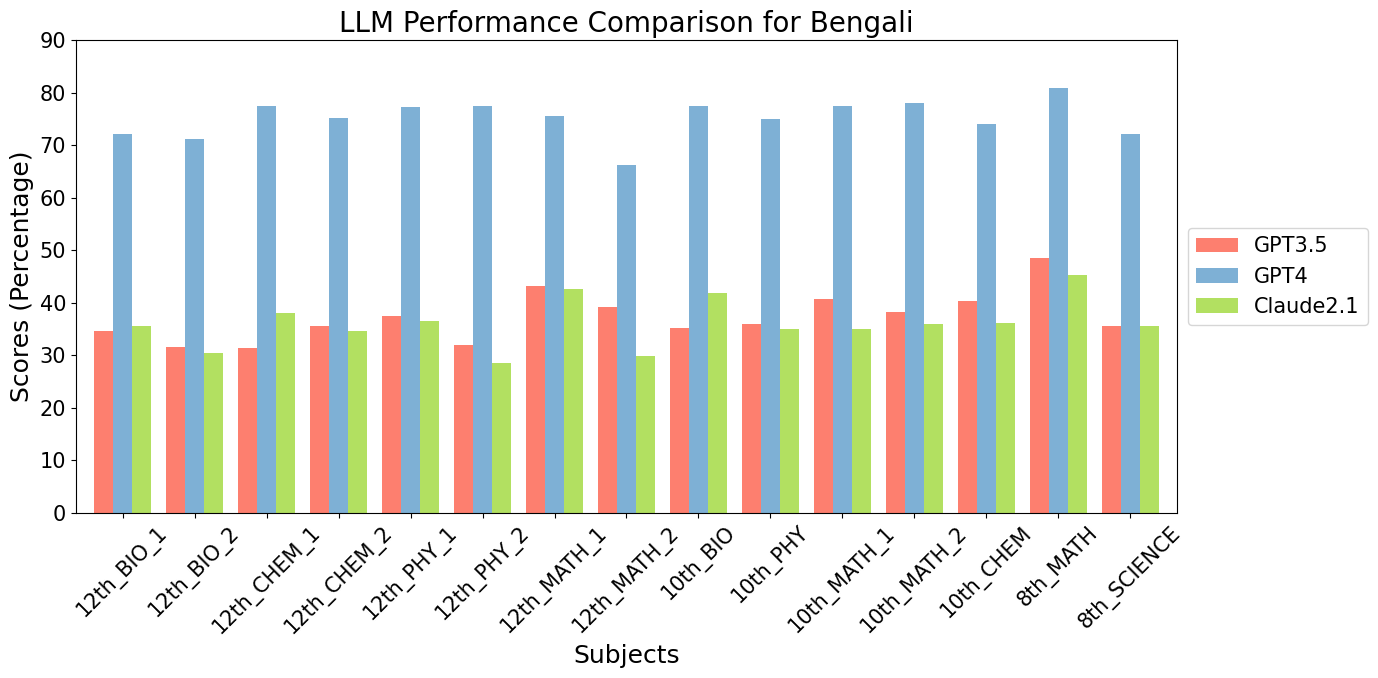}
        \caption{LLMs results in Bengali questions}
        \label{fig:all_bn}
    \end{subfigure}
    \caption{Zero-shot performance of LLMs on BEnQA dataset}
    \label{fig:all}
\end{figure*}

\begin{figure*}[t]
    \centering
    \begin{subfigure}[b]{0.49\textwidth}
        \centering
        \includegraphics[width=\textwidth]{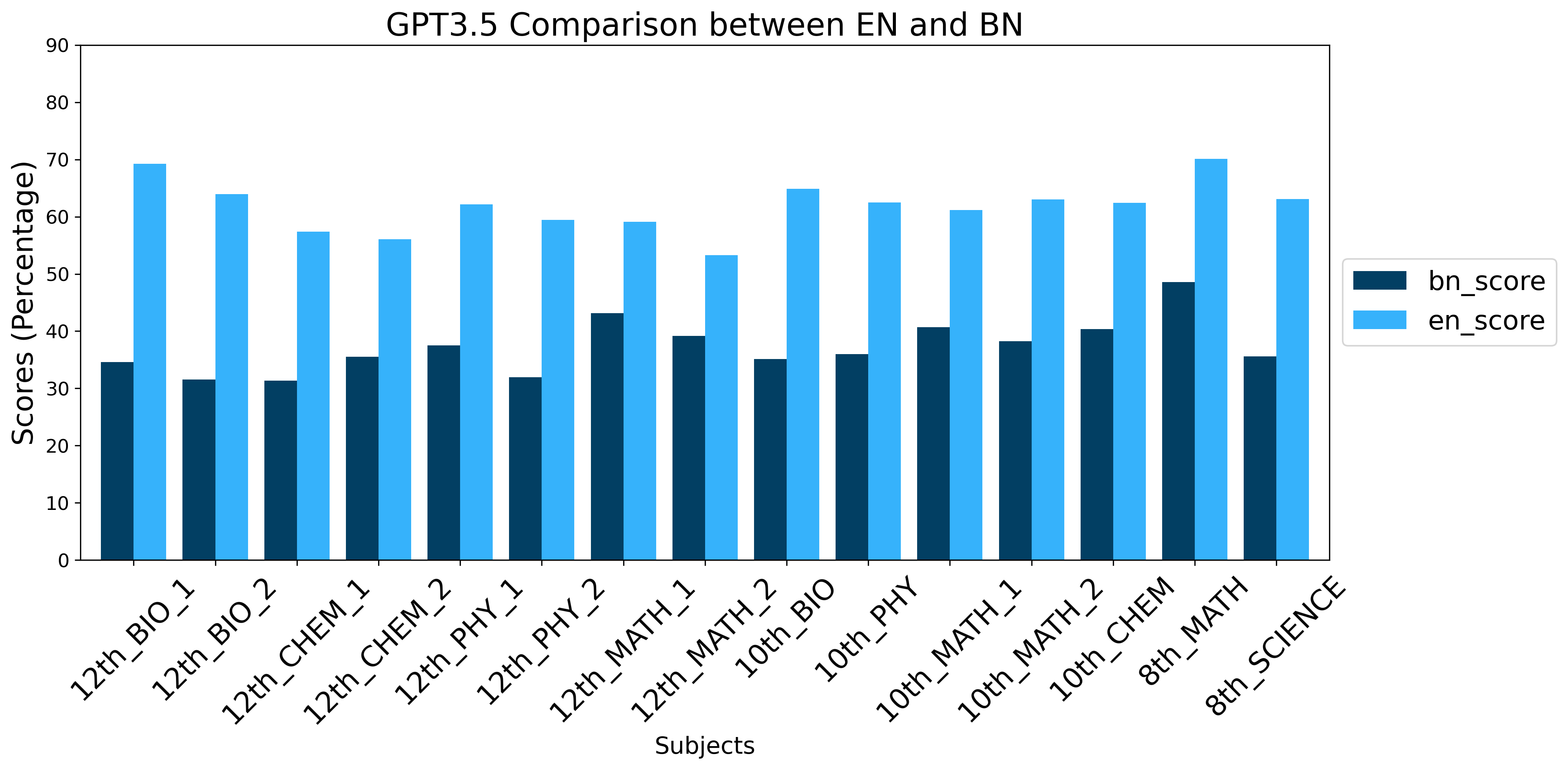}
        \caption{GPT-3.5 results}
        \label{fig:gpt3.5_en_bn}
    \end{subfigure}
    \hfill
    \begin{subfigure}[b]{0.49\textwidth}
        \centering
        \includegraphics[width=\textwidth]{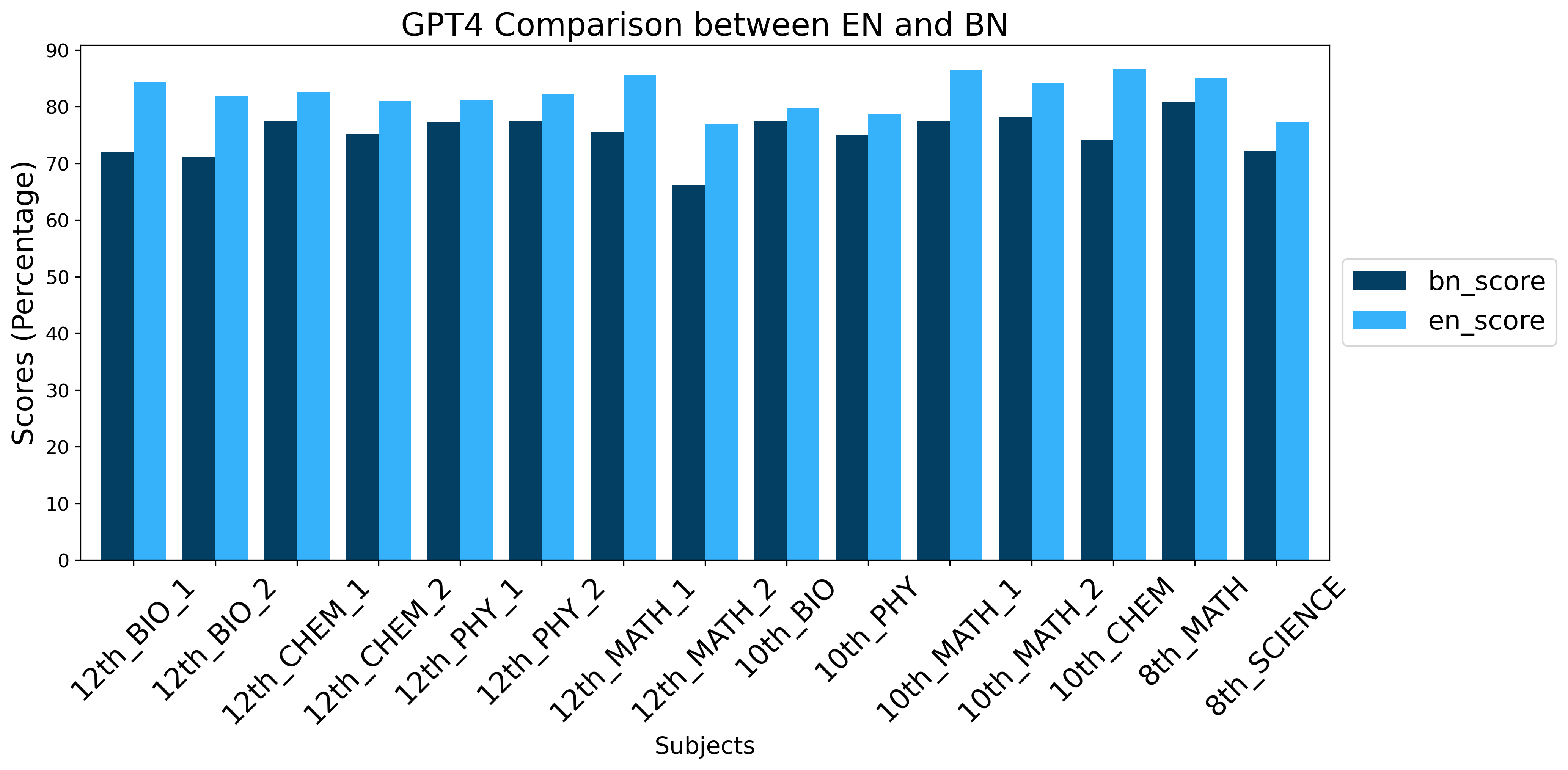}
        \caption{GPT-4 results}
        \label{fig:gpt4_en_bn}
    \end{subfigure}
    \caption{Performance comparison between Bengali and English questions}
\end{figure*}

\section{Results} % Quamran

\subsection{How do LLMs perform on BEnQA?}
% The benchmark results of all models are provided in Table \ref{tab:BEnQA_Benchmark}. 
% \input{BEnQA_table}

\paragraph{Performance across all subjects}
In this experiment, we investigate the performance of various existing LLMs across all subjects in the BEnQA dataset. The results of all models are presented in Figure \ref{fig:all}. For English questions (Figure \ref{fig:all_en}), our result shows that proprietary LLMs are far ahead of open-source LLMs such as LLaMA-2 and Mistral. Among proprietary LLMs, GPT-4 is significantly ahead, followed by GPT-3.5. We also observe that the LLMs' performance across subjects is relatively uniform, with dips in some subjects, such as 12th-grade Math II and 10th-grade Biology, indicating that these might be more challenging subjects for LLMs. The trends are also similar to Bengali questions (Figure \ref{fig:all_bn}), where GPT-4 maintains a significant score gap. Interestingly, Claude2.1 shows a much closer performance to GPT-3.5 in Bengali, a contrast to its performance in English. The full benchmark table can be found in Appendix \ref{appendix: stats}.

\paragraph{English vs. Bengali performance}
To investigate the performance disparity between English and Bengali questions, we focus on two of the highest-performing models for comparison: GPT-3.5 and GPT-4. As depicted in Figure \ref{fig:gpt3.5_en_bn}, a substantial performance gap exists between Bengali and English in GPT-3.5. However, in GPT-4 (Figure \ref{fig:gpt4_en_bn}), the gap is much smaller, and in fact, performance in both English and Bengali has improved by a large margin across all subjects. 

\subsection{Does Chain-of-Thought help?}
\label{subsec: cot}
\begin{figure}[t]
    \centering
    \resizebox{0.5\textwidth}{!}{\includegraphics{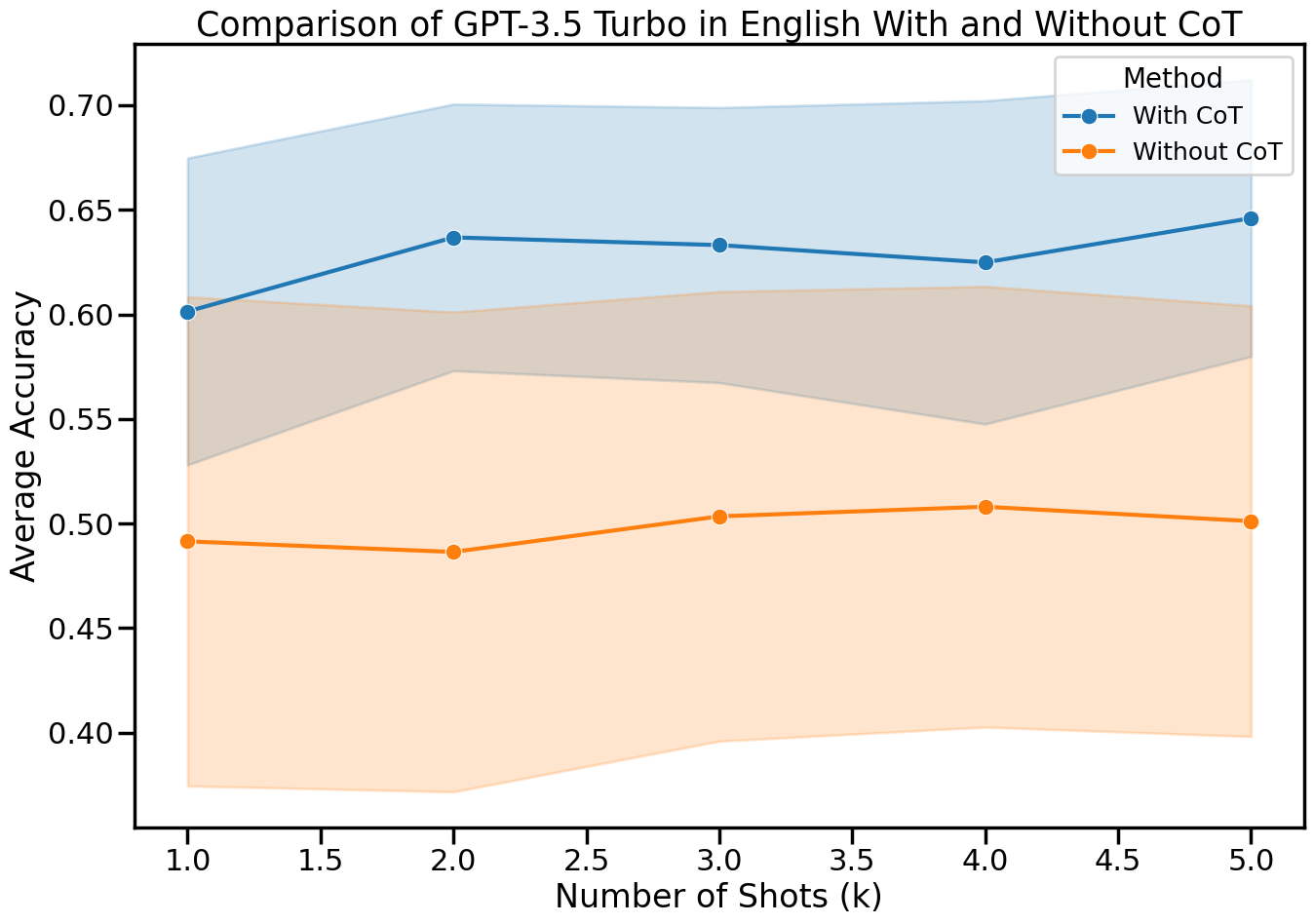}}
    \caption{Accuracy of Chain-of-Thought (CoT) reasoning in few-shot settings for GPT-3.5 Turbo in English.}
    \label{fig:cot-main_fig}
\end{figure}

\begin{figure}[t]
    \centering
    \resizebox{0.5\textwidth}{!}{\includegraphics{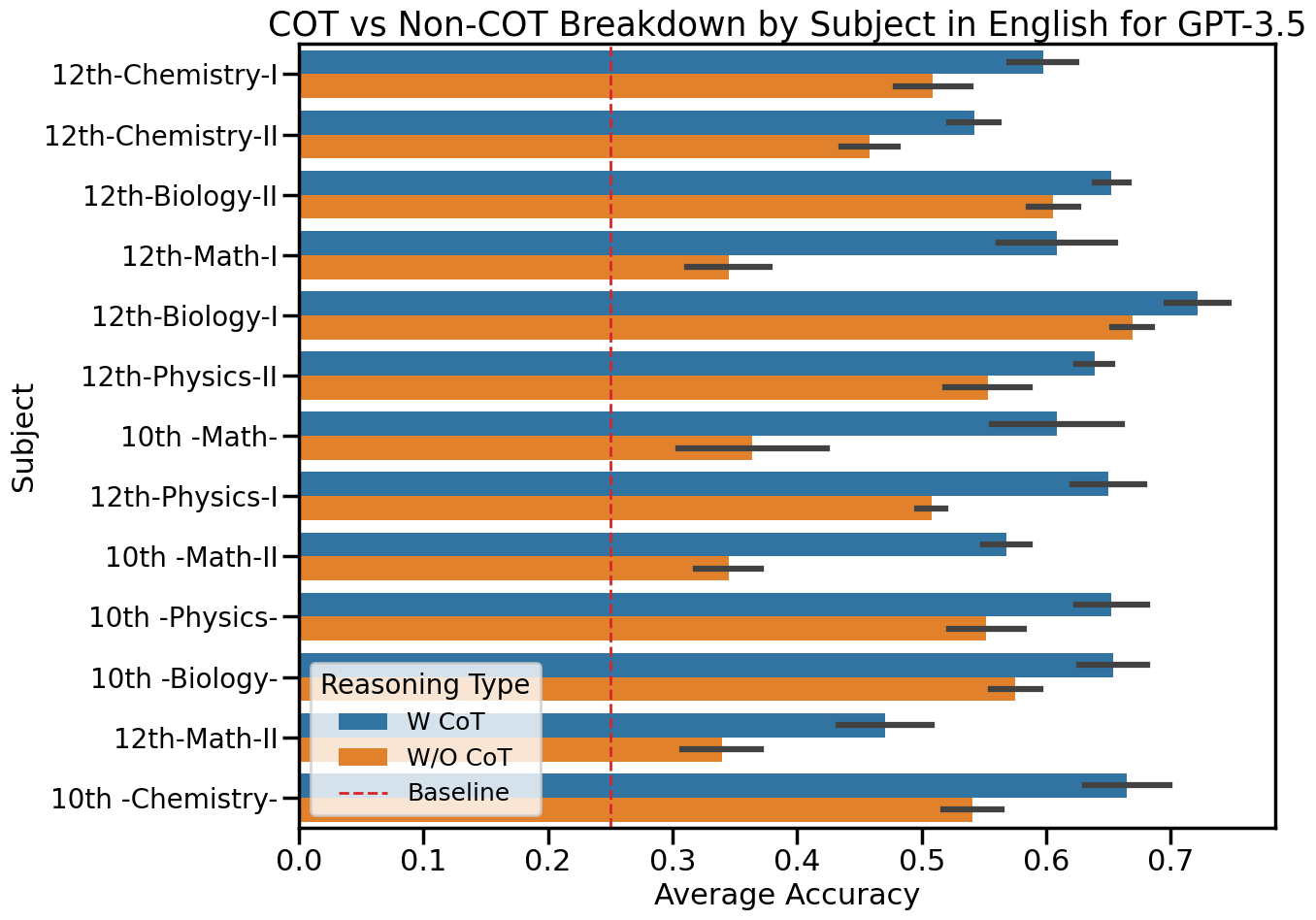}}
    \caption{Chain-of-Thought (CoT) performance broken down by subject in English. Note that some subjects benefit more from CoT than others.} 
    \label{fig:cot-subj-breakdown}
\end{figure}

\begin{figure}[t]
    \centering
    \resizebox{0.5\textwidth}{!}{\includegraphics{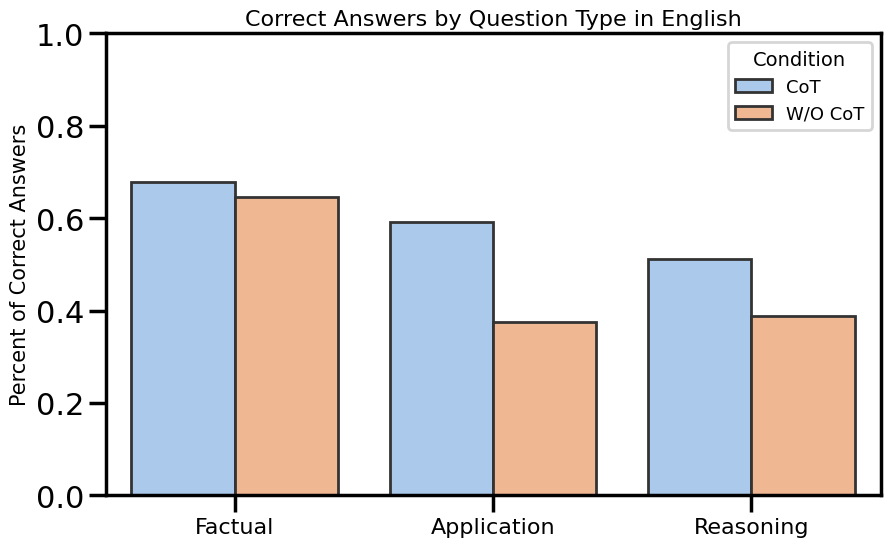}}
    \caption{Performance breakdown by question category in English.} 
    \label{fig:cot-qstype-breakdown}
\end{figure}

Chain-of-Thought (CoT) prompting is typically used to enhance the performance of LLMs in tasks that require reasoning. In order to evaluate its effectiveness on our dataset, we conduct several experiments based on subject and question type.

\paragraph{Performance by Subject}
As shown in Figure \ref{fig:cot-main_fig}, Chain-of-Thought reasoning boosts GPT-3.5 performances across the dataset in English. We also observe a similar trend with Bengali, with more details available in Appendix \ref{appendix:additional results}. Interestingly, when we decompose the result by subjects, CoT does not help all subjects uniformly as shown in Figure \ref{fig:cot-subj-breakdown}. CoT seems to help Math the most and Biology the least; this is indeed correlated with the different portion of question categories in each subject (Figure \ref{fig:sub_cat}), where we see that biology has more factual questions while math has more reasoning and application-based. 

\paragraph{Performance by Question Category}
To validate our subject-based performance findings, we further analyze the GPT-3.5 performance by question category. As depicted in Figure \ref{fig:cot-qstype-breakdown}, we indeed see that reasoning and application questions particularly benefit from CoT prompting, with accuracy improvements ranging by 10--20\%. In contrast, we observed a minimal improvement in factual questions, highlighting the need for alternative methods to enhance the factual question performance. It is also worth noting that in English, the gains in application and reasoning questions through CoT still cannot surpass the performance of factual questions. For Bengali, the improvement is mostly seen in application questions, but not so much in factual and reasoning, as elaborated in Appendix \ref{appendix:additional results}.

\begin{figure*}[t]
    \centering
    \resizebox{0.95\textwidth}{!}{\includegraphics{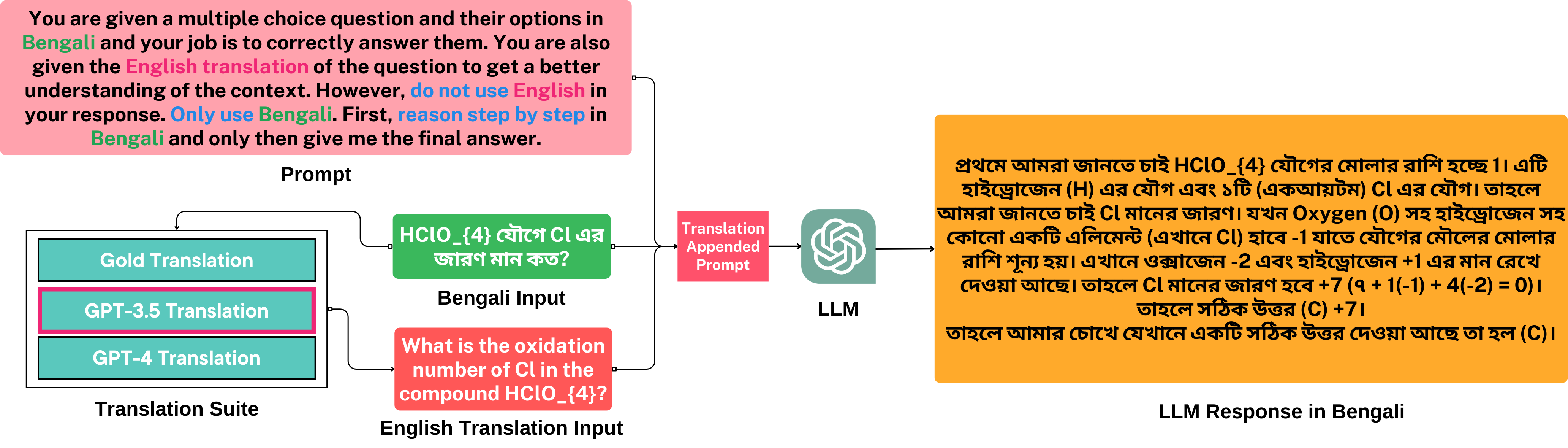}}
    \caption{Illustration of our translated appended prompting method. This is an actual example taken from our experiments. The LLM in question, GPT-3.5, answers the question correctly; however, the answer contains spelling and grammar mistakes.}
    \label{fig:prompt}
\end{figure*}

\section{Additional Experiments}
\label{sec: add_exp}
In this section, we conduct additional experiments to explore whether we can improve the performance in Bengali using better prompting. All the experiments described in this section were conducted on GPT-3.5, which provides a good value of cost-efficiency vs. capability in Bengali. 

\subsection{Can the performance be improved by appending English translation?}
\begin{figure}[t]
    \centering
    \resizebox{0.5\textwidth}{!}{\includegraphics{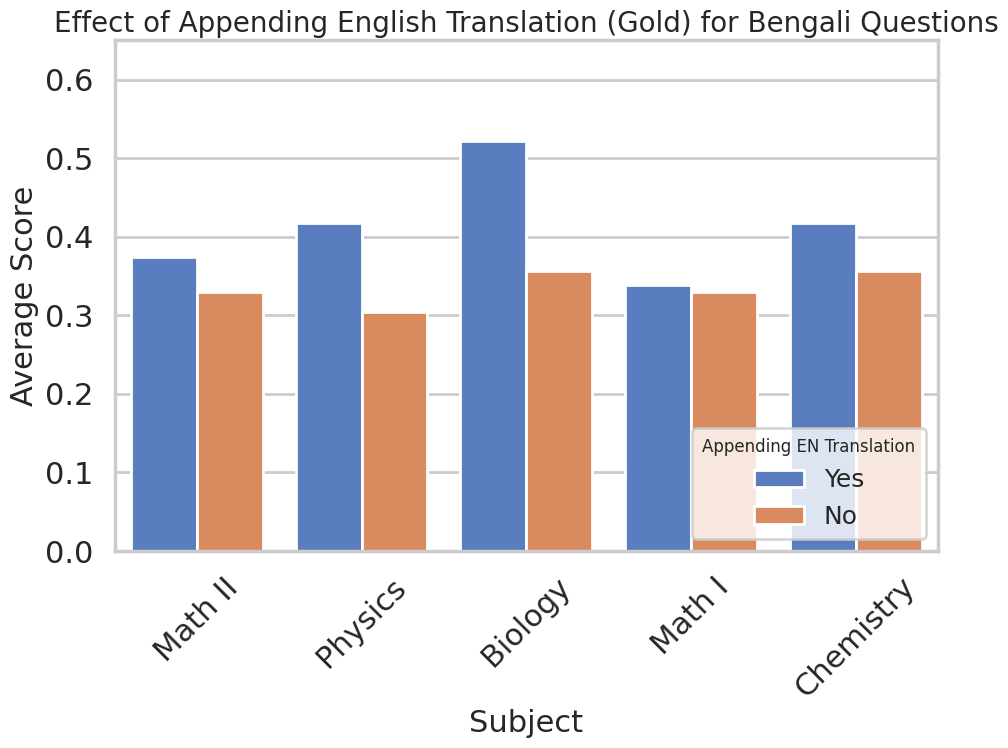}}
    \caption{Appending English translation helps to answer questions in Bengali. The model was asked to do CoT in Bengali}
    \label{fig:tr-append-exp}
\end{figure}
Our hypothesis for the models' subpar performance in Bengali might be attributed to two main factors: their unfamiliarity with Bengali scientific terminology and potential difficulty in understanding non-latin scripts \cite{lai2023chatgpt}. We expect that having access to a translation of the question would likely help the model to understand the context better. To confirm our hypothesis, we experiment with a randomly selected subset of 115 data points from each subject within the 10th-grade exam questions. We prompt the model as illustrated in Figure \ref{fig:prompt}.

% We test the models' performances in four different settings described in Section \ref{sec:exp_prompt}: baseline, appending gold translation, GPT-3.5 translation, and GPT-4 translation. The result is shown in Figure \ref{fig:tr-append-exp}.

As shown from the results in Figure \ref{fig:tr-append-exp}, appending English translation appears to have a positive impact in all subjects. The most improvement is shown in Biology, where scientific terminology can be found in the majority of questions. In contrast, for Math, the improvement is not as significant as in the other subjects. We also experimented with the case where we did not have access to gold English translations and replaced them with LLM-generated translations. Our initial experiments suggest that appending LLM-generated translations might work just as well as appending human translation. The results can be found in Appendix \ref{appendix:additional results}.

\subsection{Does the translation appended prompting method generalize to other datasets?}
\label{subsec:append_gen}
We extend our experiments to other datasets, such as COPA and Big-Bench-Hard, to evaluate the applicability of our prompting strategy across various datasets. For COPA, ask the model to answer Bengali COPA problems taken from IndicCOPA \cite{doddapaneni-etal-2023-towards}.

\paragraph{COPA}
\begin{figure}[t]
    \centering
    \resizebox{0.45\textwidth}{!}{\includegraphics{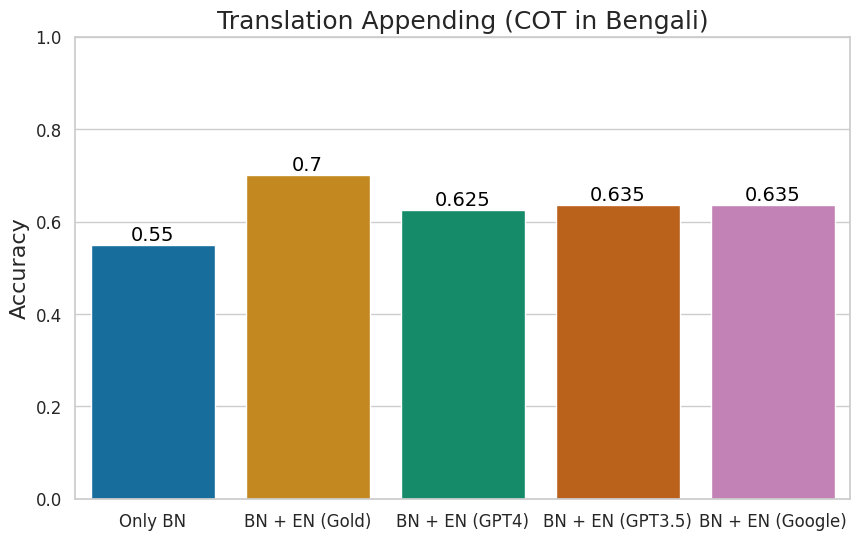}}
    \caption{Effects of our appended translation prompting method on the Bengali portion of the IndicCOPA dataset.}
    \label{fig:copa}
\end{figure}

% A noticeable improvement is observed when English translations are appended alongside Bengali, as illustrated by Figure \ref{fig:prompt}. The prompting in this case was like that of \ref{appendix: prompt}.

As shown in Figure \ref{fig:copa}, a noticeable improvement is observed in Bengali COPA when English translations are appended alongside the Bengali question. The improvement is more pronounced when we use the actual English version of the COPA dataset (marked as Gold). However, even using GPT-generated translation, accuracy improves the performance by 7 to 8\%. 

\begin{figure}[h]
    \centering
    \resizebox{0.45\textwidth}{!}{\includegraphics{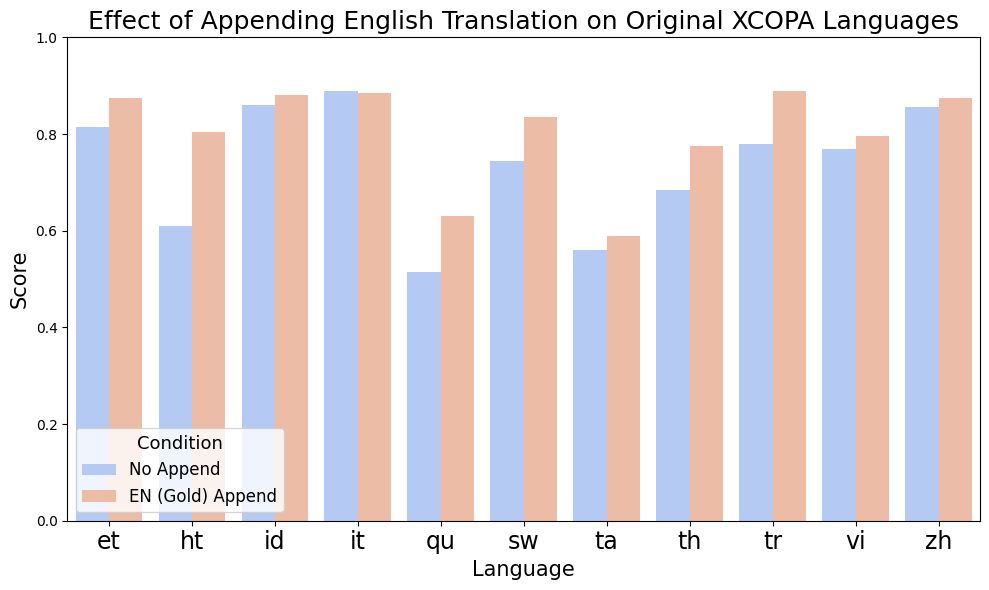}}
    \caption{Effects of our appended translation prompting method original X-COPA dataset languages. In all cases, the model was asked to do CoT in the corresponding languages.}
    \label{fig:x-copa-languages}
\end{figure}

We also try to see if translation appended prompting can be helpful for the other languages included in the original X-COPA \cite{ponti-etal-2020-xcopa} languages. The results are shown in the Figure \ref{fig:x-copa-languages}. Appending English translation increases accuracy in all but one language, with the effects most pronounced in low-resource languages. Further research in utilizing this prompting technique is an interesting direction left for future work. 

\paragraph{Big-Bench-Hard Bengali }
For this experiment, we curate Big-Bench-Hard Bengali dataset using GPT-4. Two native Bengali speakers generated prompts for each Big-Bench task iteratively and then generated the prompt. 
% As seen in Figure \ref{fig:gpt 3.5 tr_append}, English consistently outperforms Bengali across all tasks in BIG-Bench Hard. 

Our experiment shows that on average, there is a 5.8\% increase in BBH-BN performance in Bengali when we append gold English translation, while appending GPT-4-generated English translation provides an average of 4.2\% improvement. Details about Big-Bench-Hard task selections and results can be found in Appendix \ref{appendix: extra_experiments}.

\section{Discussion}
Unlike most other bilingual or multilingual NLP datasets, which are usually first curated in English and then translated into other languages, questions in our BEnQA datasets are curated in Bengali first and then translated into English by school teachers. This approach could also provide an interesting outlook on how L2 English speakers might use English and whether LLMs are robust to such linguistic variations. 

Besides, our benchmark results highlight that LLMs, such as ChatGPT, still have some way to go in terms of catching up with performances in low-resource languages like Bengali. One interesting aspect we noticed is that, in Bengali, it is much harder to make the model output in a predefined format for automated evaluation. Future research should focus on improving the instruction-following ability of multilingual models.

% the model sometimes reaches answer directly without doing Chain-of-Thought reasoning even though it was asked to do chain of thought reasoning: suggesting that LLMs are probably not very good at instruction following in non-English languages compared to English.

Another key finding is the substantial performance gap between open-source and proprietary language models. The open-source models, which are more accessible in the context of developing countries, must stride ahead to catch up with the proprietary models to make sure the benefits of AI, particularly LLMs, are not limited to only some demographics. A recently published open multilingual LLM, Aya \cite{ustun2024aya}, represents a promising step in this direction.

Furthermore, we demonstrated the feasibility of having LLMs respond in the target language, while leveraging the advantage of high-resource languages, such as English, in the inference pipeline by utilizing translation of the query. The benefit of this approach is that we do not need to have access to the manually-created gold translation (as we had in this case): translation using the same model or more powerful/specialized models would work too. However, the choice of translation method, whether it is another more proficient LLM or a domain-specific fine-tuned translation model, can influence the performance. This finding necessitates further research to optimize and refine the use of LLMs, especially in the context of low-resource languages.

\section{Conclusion and Future Work} % Quamran
In this paper, we introduced BEnQA, a locally sourced dataset from Bangladesh, comprising middle-school and high-school level exam questions in both English and Bengali. Our dataset is provided in parallel format, allowing us to benchmark the performance discrepancy between both languages in a fairer way. Upon benchmarking several Large Language Models (LLMs) with this dataset, we found that performance in Bengali lags behind English, even for the best-performing models; and current open-source models significantly lag behind proprietary models. We also explored whether the performance in Bengali questions can be improved by appending English translations to the prompts. This approach resulted in performance enhancements across most subjects in the BEnQA dataset on GPT-3.5 model. Notably, this improvement also extends to other datasets, such as COPA and Big-Bench-Hard. These findings pave a promising direction for future research in improving the LLMs performance in low-resource languages, specifically in Bengali.

Our findings open up several promising directions for future research. Our results have highlighted a possible gap in the current models' understanding of Bengali terminology, leading us to develop a very simple prompting method that helps the performance in Bengali data significantly. We also show the efficacy of this prompting approach on Bengali COPA, X-COPA, and Bengali Big-Bench-Hard dataset. Future research should also focus on to see to what extent such prompting methods utilizing high-resource languages generalize to other datasets and languages.

% This approach has the potential to enhance the models' proficiency in handling the linguistic nuances and specific terminology in the Bengali language, especially within the context of science questions.

\section*{Limitation} % Sheikh
Our works have several limitations that should be acknowledged. Firstly, our dataset primarily focused on text-based questions, as we discarded figure-based questions during the dataset curation. This limitation might restrict the scope of our findings, as questions with visual elements often require more reasoning steps. Additionally, since the questions are multiple-choice, there might be a possibility that the models use a shortcut to answer the questions, especially for questions that do not require advanced reasoning skills like factual questions. Despite these limitations, our dataset serves as an important starting point for benchmarking LLMs in Bengali, which currently have limited resources available for reasoning and science question-answering.

\section*{Ethical Considerations}
The exam questions in BEnQA dataset are freely available and have been manually curated and reviewed to minimize the presence of harmful content. This dataset will be publicly accessible and distributed under the CC BY-SA 4.0 license. Our work has been reviewed and received approval from the Institutional Review Board (IRB) at our institution. All annotators involved in this project were compensated above the minimum wage.

% \section*{Acknowledgements}

\bibliography{anthology,custom}

\appendix

% \section{Dataset Curation Heuristics}
% \label{appendix:dataset_curation}
% Even though the national board exams of Bangladesh are held in both English and Bengali versions, the question index in each version is not necessarily the same (i.e., the first question of the Bengali version can be the eleventh question in the English version), which is why we developed a very simple, yet effective algorithm for aligning the English and Bengali questions. 

% First, we translated each Bengali question to English via Google Translate API and then within each subject we matched the Google translation with actual English translation of the question using cosine similarity of their embeddings. We used OpenAI \texttt{text-embedding-ada-002} embedding to achieve this. After the initial matching, two native Bengali speakers with proficiency in English verified each question manually to validate whether each Bengali question belonged to the same English question.

% \begin{figure*}[h]
%     \centering
%     \resizebox{0.95\textwidth}{!}{\includegraphics{Figures/parallel_corpus_making.png}}
%     \caption{BEnQA Parallel Corpus Making Scheme}
%     \label{fig:corpus_making}
% \end{figure*}

% We also filter out a few questions for which the ground truth was different in English and Bengali. Manual inspection revealed that such cases usually correspond to errors on behalf of the annotators or the questions themselves. 

\section{Effect of Grammar Mistakes in BEnQA}
\label{appendix:grammar_mistakes}
\begin{figure}[t]
    \centering
    \resizebox{0.5\textwidth}{!}{\includegraphics{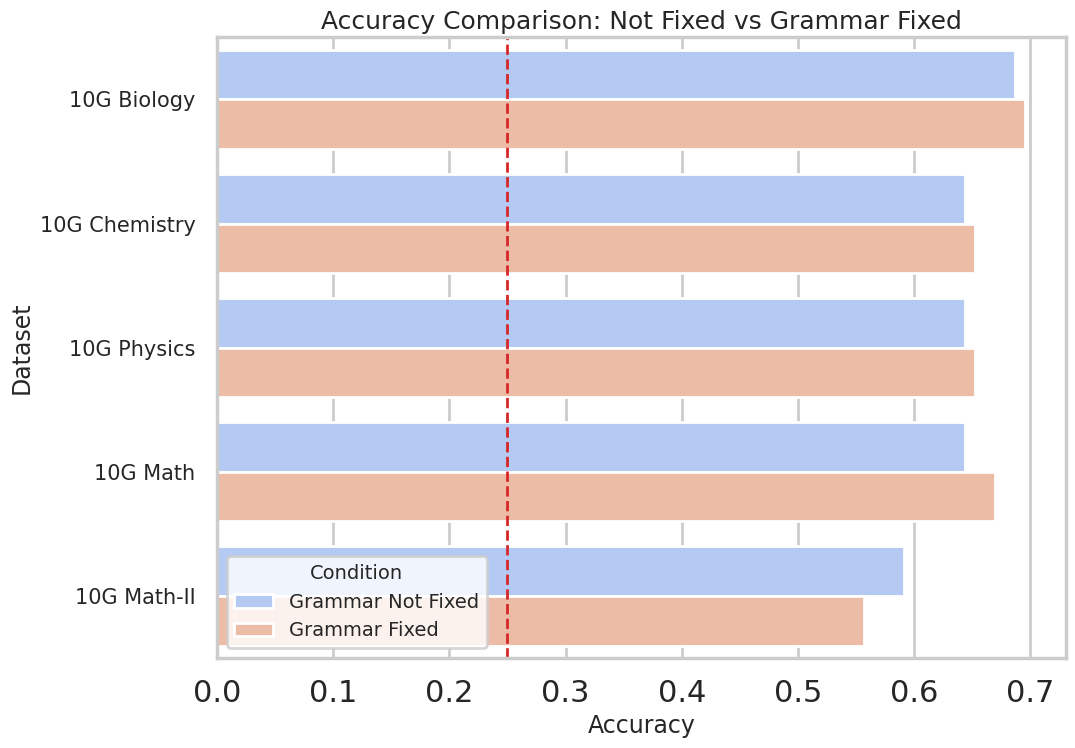}}
    \caption{Effects of grammar mistakes on GPT-3.5 performance.}
    \label{fig:grammar-fix}
\end{figure}

To see the effect of subtle grammar mistakes and English translations that sound unnatural to native speakers, we first take 120 questions of each subject of the 10th-grade exam and prompt GPT-4 to fix grammatical issues and unnatural translations if there are any. One native Bengali speaker proficient in English then went through the results and made necessary adjustments to the GPT-4 fixed version.\footnote{oftentimes GPT-4 destroys the original meaning of the question when doing grammar fixing, which is why we did not use this protocol to fix grammar for the whole dataset} The performance of GPT-3.5 on this subset of the dataset is presented in Figure \ref{fig:grammar-fix}.
The difference between a grammar mistake-free version and the version with grammar mistakes is very little, and a manual inspection revealed the slight discrepancy that is shown in Figure \ref{fig:grammar-fix} emerged from the stochastic nature of GPT-3.5 rather than any grammar mistakes.

\section{Additional BEnQA Results}
\label{appendix:additional results}

\subsection{Few Shot Results}

\begin{figure}
    \centering
    \resizebox{0.5\textwidth}{!}{\includegraphics{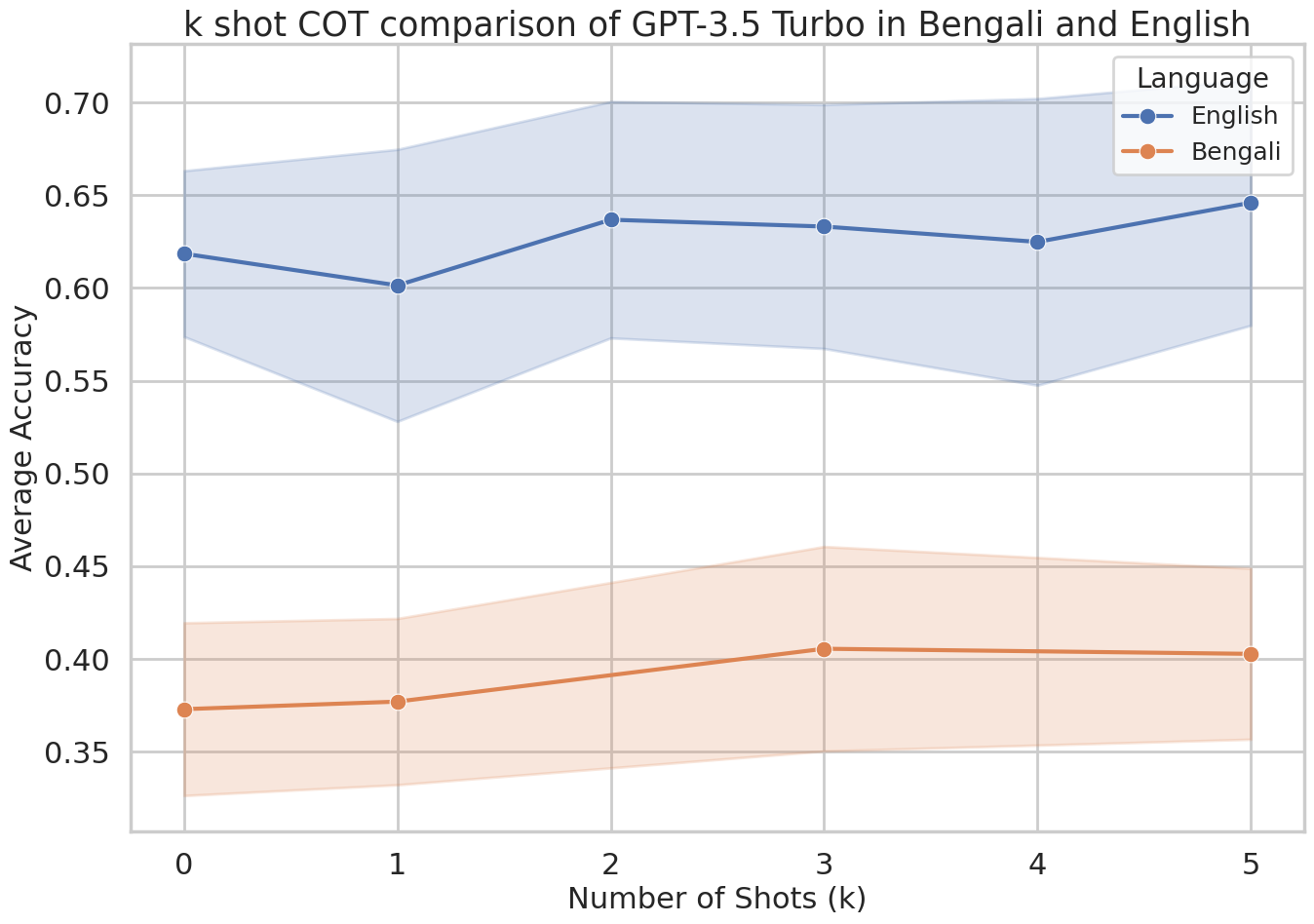}}
    \caption{k-shot prompting in Bengali and English on GPT-3.5. Note that zero shot prompting had more detailed system prompt than for few shot prompting (See Appendix \ref{appendix: prompt})}
    \label{fig:k shot}
\end{figure}

We adhere to zero shot prompting throughout the paper as the preliminary experiments revealed that the difference between zero shot and few shot settings is not very significant (Figure \ref{fig:k shot}) and conducting extensive few shot experiments is costly for proprietary models, and even more so in Bengali due to infertile tokenization. 

\paragraph{Few-Shot Example Preparation:}
When preparing the few shot examples, it was made sure that at least the first three examples cover our categorization (factual, procedure and knowledge, reasoning) as described in Section \ref{sec:data_prop}. The five examples also covered all types of questions that (multiple choice and multi-selection type) usually appear in the exams. Finally, the examples in each topic were topic-wise diverse for each subject.

\subsection{Effect of Chain of Thought Prompting}
\begin{figure}
    \centering
    \resizebox{0.5\textwidth}{!}{\includegraphics{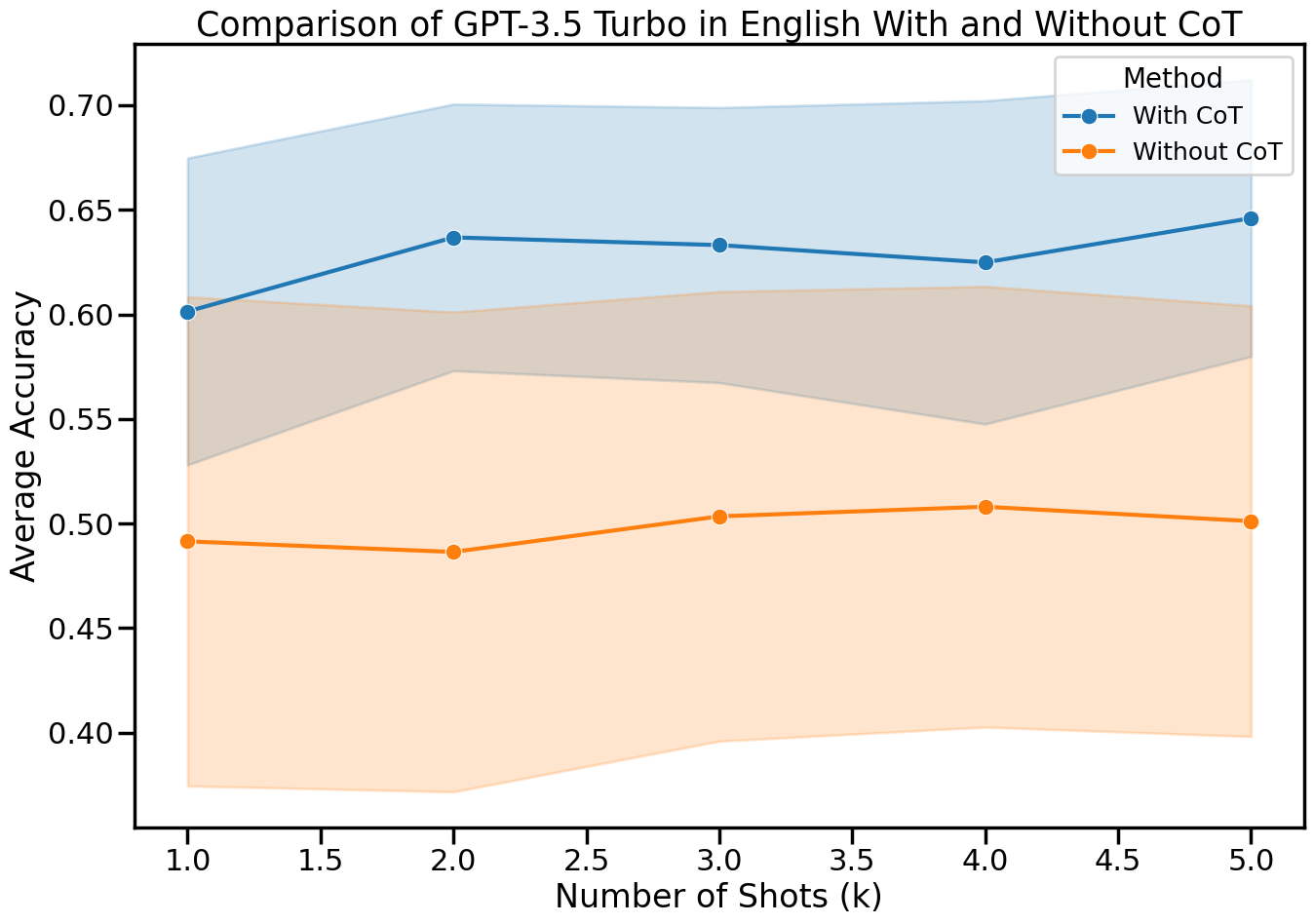}}
    \caption{Effect of Chain of Thought reasoning across k shot on GPT-3.5 for BEnQA English}  
    \label{fig:cot-kshot}
\end{figure}

Chain of Thought performance for English in 1 to 5-shot settings is given in Figure \ref{fig:cot-kshot}. In all cases doing CoT improves performance. 

\begin{figure}
    \centering
    \resizebox{0.5\textwidth}{!}
    {\includegraphics{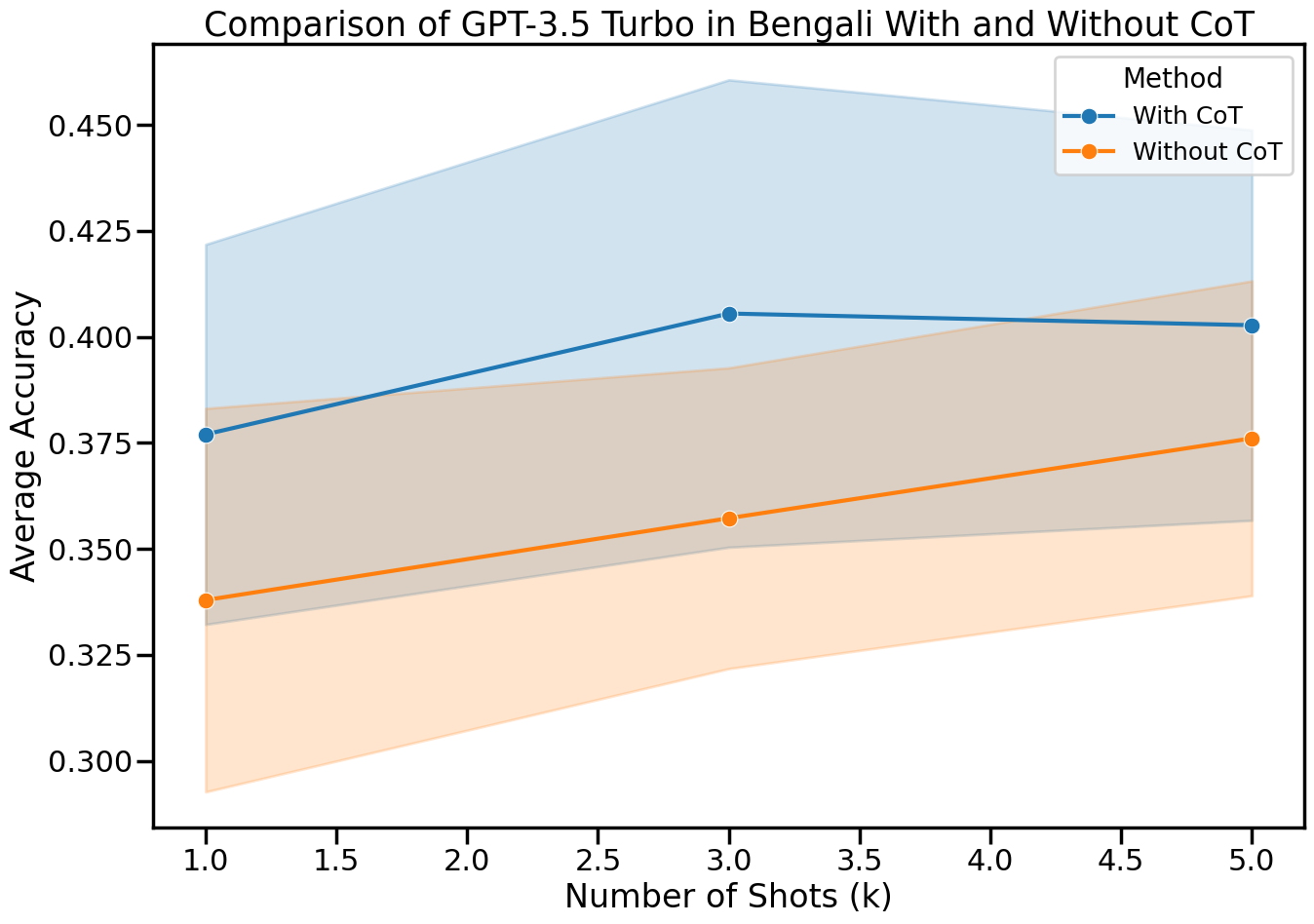}}
    \caption{Effect of CoT reasoning across k shot on GPT-3.5. Note that in this Figure k is 1, 3 and 5}
    \label{figure: bn-cot-k-shot}
\end{figure}

Similar observation can be made for GPT-3.5 in Bengali as well (Figure \ref{figure: bn-cot-k-shot}). We only chose k = 1, 3, and 5 to prevent high API costs associated with low-resource languages.
%Open-source models benefit from CoT (Figure \ref{fig:llama13-b ssc cot}).

\begin{figure}[h]
    \centering
    \resizebox{0.5\textwidth}{!}{\includegraphics{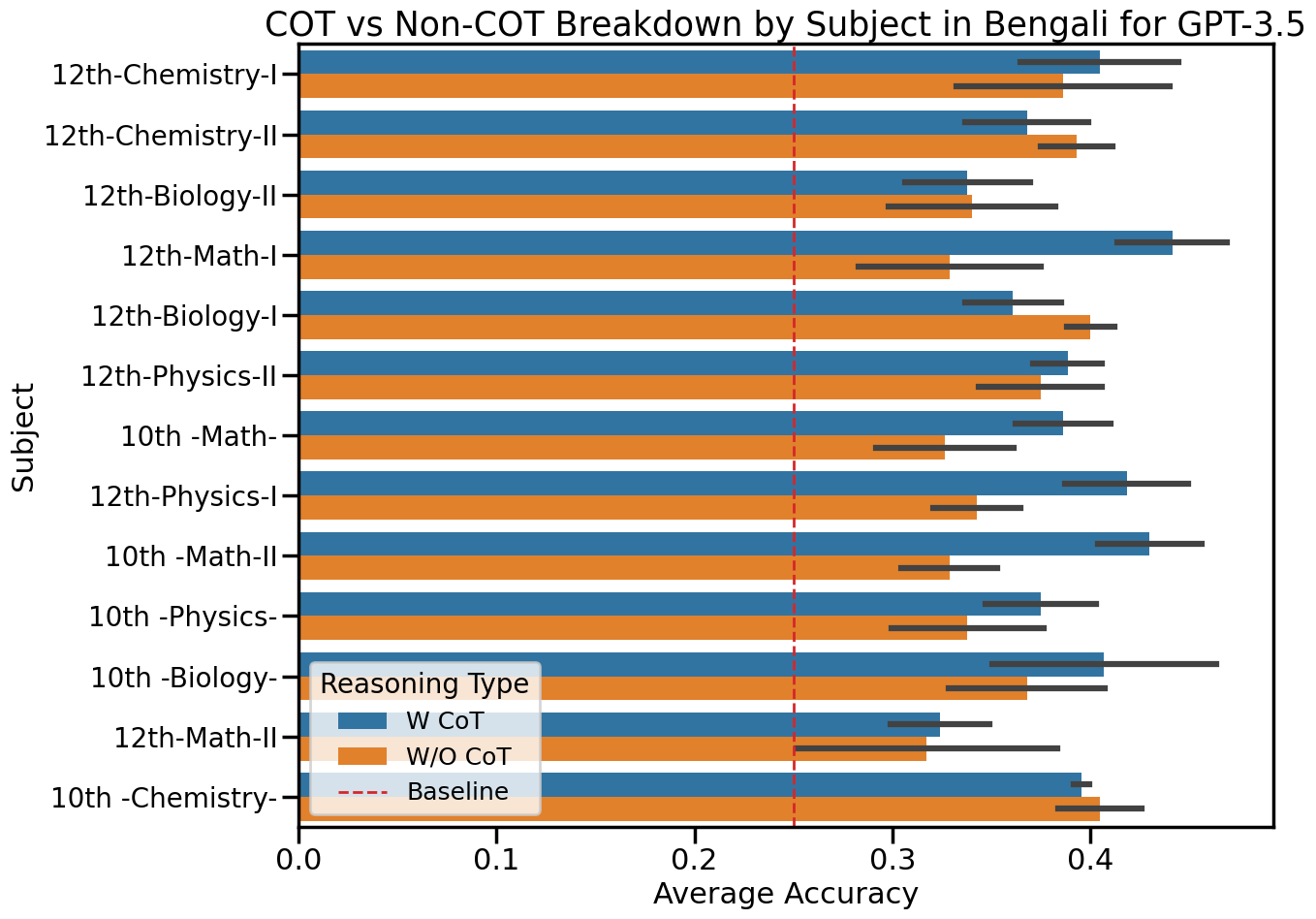}}
    \caption{CoT performance breakdown by subject in Bengali.}  
    \label{fig:bn-cot-breakdown-by-subj}
\end{figure}

\paragraph{CoT Subjectwise Breakdown:}
Similar to what we have observed in Section \ref{subsec: cot} We also note that doing CoT improves performance for Bengali. Figure \ref{fig:bn-cot-breakdown-by-subj} shows the subjectwise CoT performance breakdown for GPT-3.5.

\begin{figure}[h]
    \centering
    \resizebox{0.5\textwidth}{!}{\includegraphics{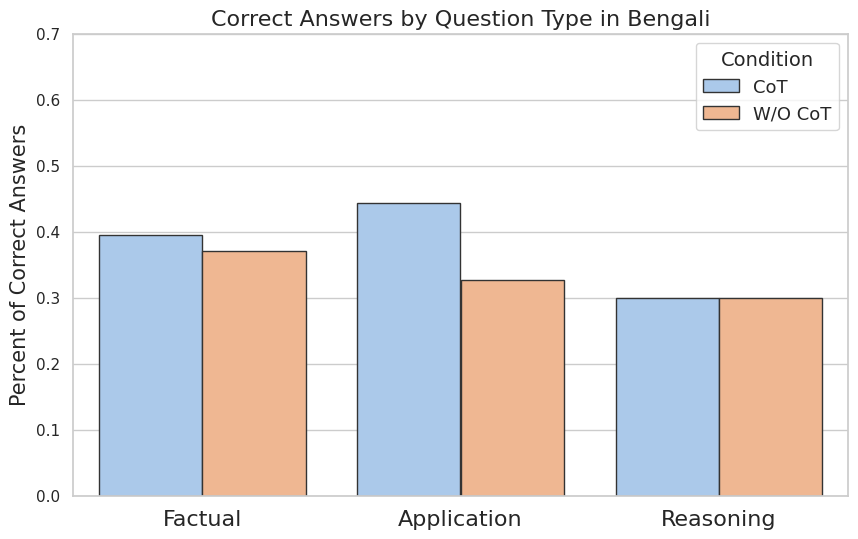}}
    \caption{CoT performance breakdown by question type in Bengali.}  
    \label{fig:bn-cot-breakdown-by-type}
\end{figure}

We can also see the CoT performance breakdown by question type for Bengali in Figure \ref{fig:bn-cot-breakdown-by-type}.

% Open source results commented out
% \begin{figure}[h]
%     \centering
%     \resizebox{0.5\textwidth}{!}{\includegraphics{new_figures/llama13b-cot-vs-no-cot.png}}
%     \caption{Subject breakdown of LLama-13B model CoT performance}  
%     \label{fig:llama13-b ssc cot}
% \end{figure}
% Figure \ref{fig:llama13-b ssc cot} shows the subjectwise breakdown for CoT performance for 10th grade subjects for open source llama models.
\subsection{Does LLM-generated translation work as well for appending?}

One obvious question our translation appending prompt method raises is that in most cases we do not have access to human generated translation to append with the model input. In that case, does machine translation work as a replacement?

\begin{figure}[h]
    \centering
    \resizebox{0.5\textwidth}{!}{\includegraphics{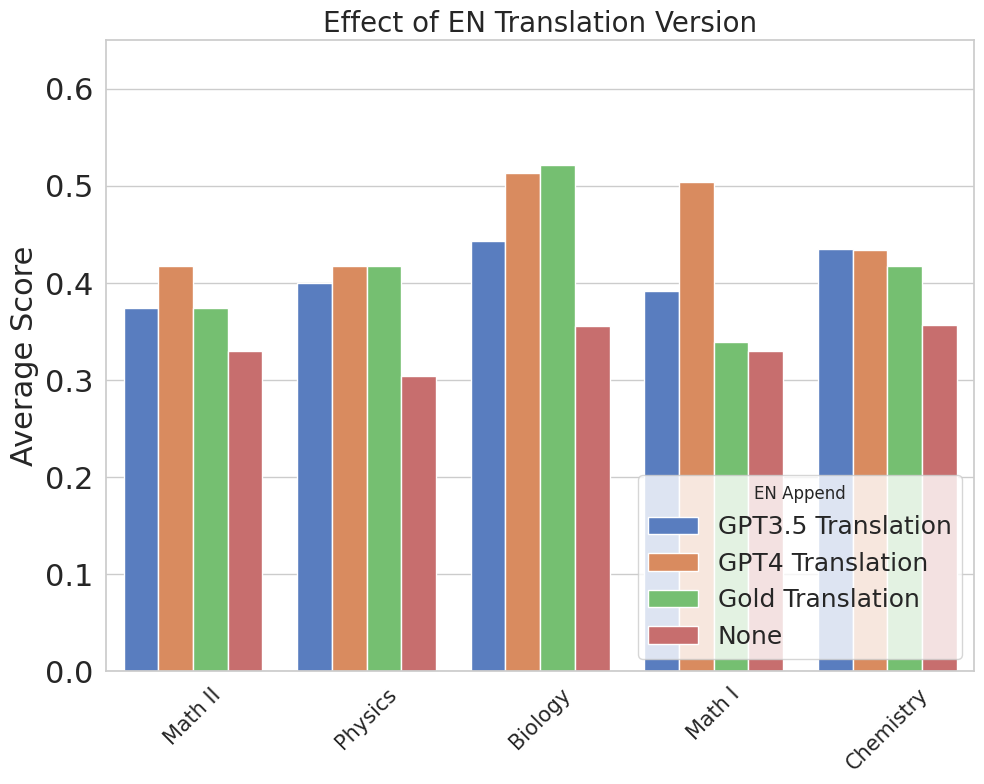}}
    \caption{Effect of appending LLM generated English translation in appending experiment. \texttt{None} refers to the case where no English translation was appended.}  
    \label{fig:ssc-append-tr-quality}
\end{figure}

Figure \ref{fig:ssc-append-tr-quality} shows that appending LLM-generated English translation works just as well, in fact, it outperformed appending original translation for some of the subjects. We only used GPT-3.5 and GPT-4 translations in this experiment and did not incorporate Google Translate; because, in our observations, equations written in \LaTeX format tend to get broken by Google Translation.

The result of this experiment leaves open the possibility of incorporating such prompting techniques for a wide range of multilingual tasks where obtaining manual English translation would be quite difficult.

\section{Evaluation of Translation Quality Using Backtranslation}
\label{appendix: translation_quality}
\begin{figure}
    \centering
    \resizebox{0.5\textwidth}{!}{\includegraphics{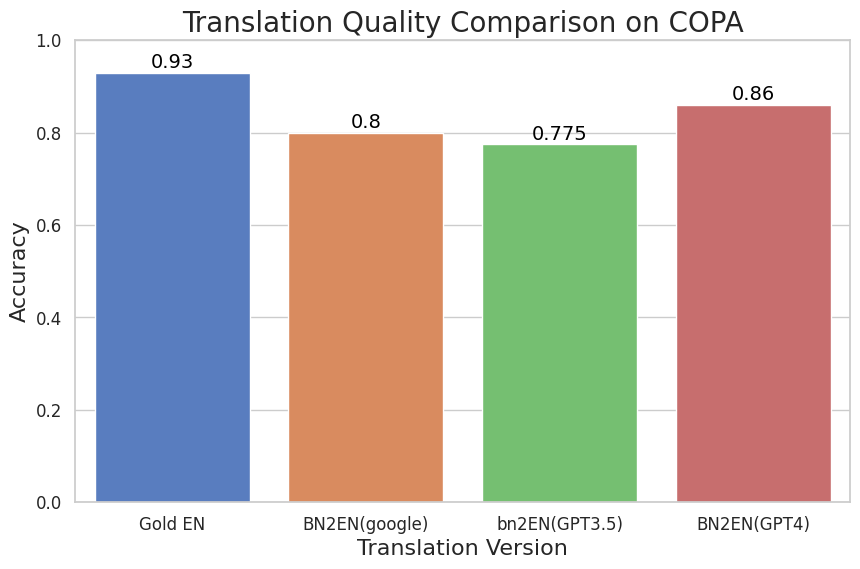}}
    \caption{Backtranslating COPA from Bengali to English and seeing their performance. Gold EN refers to the original English COPA dataset}
    \label{fig:copa-translation-quality}
\end{figure}

We use the COPA dataset to validate the translation quality of different models. We first take the Bengali translation of COPA dataset from \cite{doddapaneni-etal-2023-towards} and backtranslate the Bengali translation of COPA dataset from Bengali to English using GPT-4, GPT-3.5 and Google Translate and then compare it with the performance on original English COPA. The results are in the Figure \ref{fig:copa-translation-quality}.

As we can see, GPT-4 translation (BN to EN) provides the closest results to the original English translation, while GPT-3.5 and Goolgle Translate are slightly lower.

\section{Additional Experiments}
\label{appendix: extra_experiments}
The detailed setup for COPA and Big-Bench experiments was like this:
\begin{itemize}
    \item \textbf{EN:} Asking only in English (using the original data). 
    \item \textbf{BN:} Asking only in Bengali.
    \item \textbf{BN + EN (Gold):} Asking in Bengali and appending the original data. 
    \item \textbf{BN + EN (GPT-3.5):} Asking in Bengali and appending the English translation done by GPT-3.5.
    \item \textbf{BN + EN (GPT-4):} Asking in Bengali and appending the English translation done by GPT-4.
\end{itemize}

\subsection*{Big-Bench Hard Experiments}
\paragraph{Big Bench Hard Translation: }From the 23 datasets officially provided by BIG-Bench Hard, we selected 14
tasks that can have equivalence in Bengali\footnote{some BBH tasks like manipulating English alphabet letters, which do not have a direct `translation' in Bengali}. We chose Causal Judgement, Date Understanding, Disambiguation QA, Formal Fallacies, Logical Deductions Five, Logical Deductions Seven, Logical Deductions Three, Multistep Arithmetic, Navigate, Object Counting, Penguin In a Table, Reasoning About Colored Objects, Temporal Sequences, and Web of Lies tasks for our work.

These 14 tasks were machine-translated into Bengali using GPT-4 with three human-annotated examples as prompts for each task. The prompts were made iteratively by two Bengali speakers to reflect the nature of each task. 

\begin{figure*}[h]
    \centering
    \resizebox{0.95\textwidth}{!}{\includegraphics{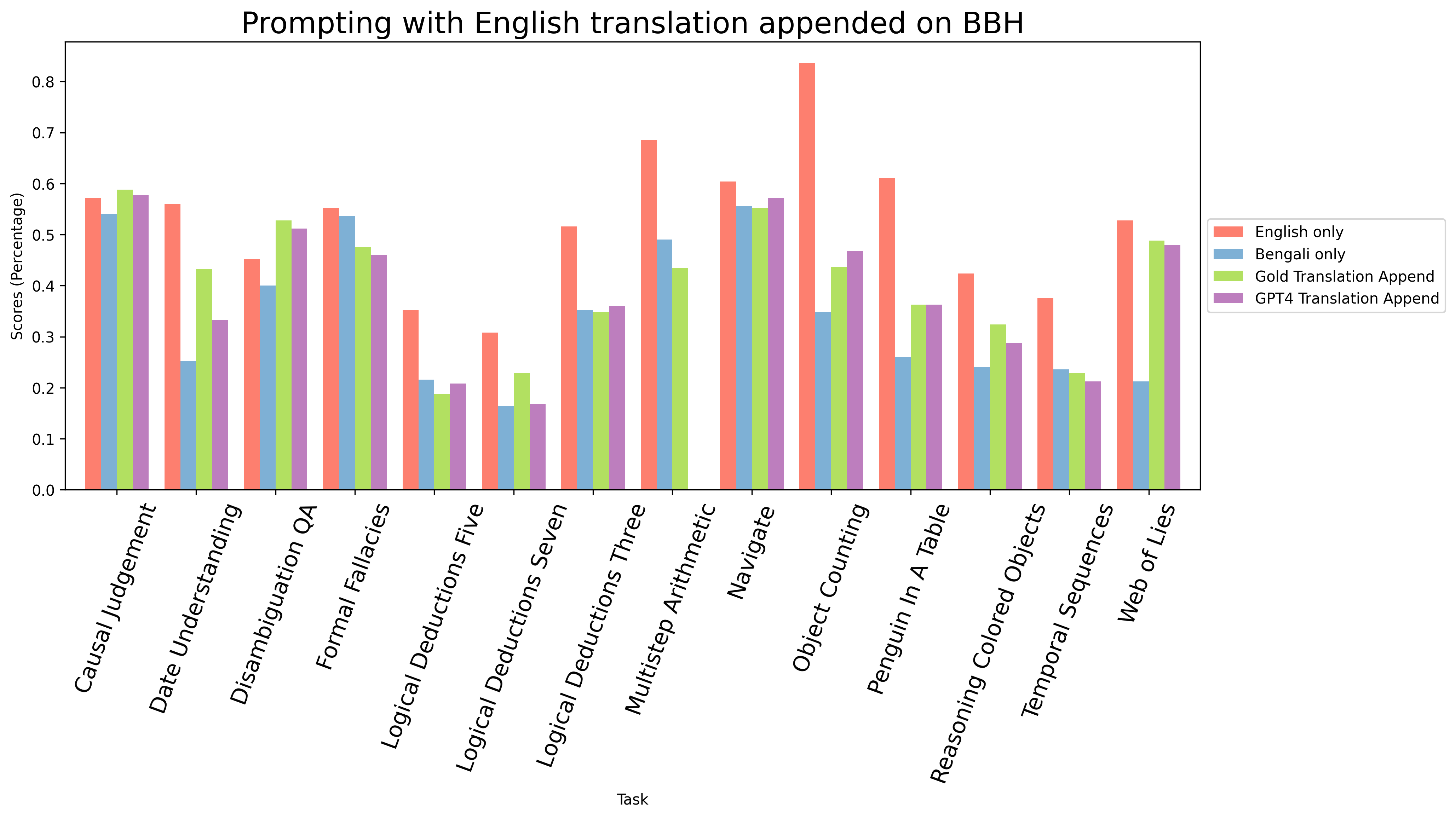}}
    \caption{Effects of our translation appended prompting method on BIG-Bench Hard dataset. Here Gold Translation Append refers to the case where the question was asked in Bengali and the original English translation was appended, while GPT-4 Translation Append refers to the case where the English translation was generated using GPT-4}
    \label{fig:bbh-tr-append}
\end{figure*}

Figure \ref{fig:bbh-tr-append} shows the results for appending English translations in Big-Bench-Hard experiments.

In 7 out of 14 cases, appending English translation helped, while in three cases, it slightly hurt the performance. For the remaining four cases, performance remained largely unaffected.  The detailed numbers can be found in Table \ref{tab:BBH_benchmark}.

\section{Prompts Used for Experiments}
\label{appendix: prompt}
This section contains all the original prompts used for all the experiments. 

\paragraph{Grammar-Corrected BEnQA Question Preparation: }
We use the prompt given in Table \ref{tab:grammar_fix} to fix the grammatical issues of the original questions in English with GPT-4.
\begin{table*}[hp] 
    \centering 
    \fbox{
        \parbox{\textwidth}{ 
            \textit{
  You are given a multiple choice exam question in English (along with their choices). \\ \\
  The question is mostly good, but sometimes contains minor grammatical mistakes and non-standard vocabulary. Your job is to make it sound natural and fluent. \\
  You are also given the choices of the question to get a better understanding of the context. \\
  However, do not return those choices in your response. Only return the question. Make sure you do the full question, not just the part of it. \\
  If there is any equation or formula in the question, do not modify them. \\ \\
  The question to be fixed is given below: \\
            }
        }
    }
    \caption{Prompt for Making Grammar-Corrected Questions by GPT-4}
    \label{tab:grammar_fix}
\end{table*}

\paragraph{BEnQA Dataset Categorization: }
We mentioned in \ref{sec:data_prop} that we classified our dataset questions into these three categories: Factual Knowledge, Procedural \& Application, and Reasoning using GPT-4. We use the prompt given in Table \ref{tab:categorizer} to categorize them.

\begin{table*}[ht] 
    \centering 
    \fbox{
        \parbox{\textwidth}{ 
            \textit{
                You are given a multiple choice question and your job is to tell what kind of reasoning is required to solve the problem.\\ \\
                Choose from the following options:\\
                1. Factual Knowledge: The question requires only the knowledge of basic facts, dates, events, concepts, etc.\\
                2. Procedural and Application: The question requires the ability to apply a procedure or a formula to solve the problem.\\
                3. Reasoning: The question requires the ability to do multistep reasoning to solve the problem.\\ \\
                The question is given below:
            }
        }
    }
    \caption{Prompt used to categorize questions in BEnQA dataset.}
    \label{tab:categorizer}
\end{table*}

For each question in the dataset, we take zero-shot approach with the prompt for categorization. 

Table \ref{category_Table} represents the subject-wise question category in tabular visualization.

\paragraph{Chain-of-Thought Prompting for BEnQA Zero-shot benchmark: }
Table \ref{tab:zero-shot_prompt} shows the prompt for zero-shot benchmarking with the Proprietary models.

\begin{table*}[htbp]
    \centering
    \fbox{
        \parbox{\textwidth}{ 
            \textit{
                You are given a multiple choice question and their options in English/Bengali and your job is to correctly answer the question. First reason step by step in English/Bengali and only then give me the final answer as "a", "b", "c" or "d".\\ \\
                Keep these in mind:\\
                1. Only include the letter a, b, c, d as your final answer. Do not include the option text.\\
                2. Every question will have an answer in the given options. So, DO NOT say that none of the answers are correct.\\
                3. ONLY ONE of the given options will have the answer. So DO NOT provide multiple options as answers.\\
                4. The questions contain enough information to solve the problem, so DO NOT say that you need additional information.\\ \\
                The question is given below:
            }
        }
    }
    \caption{Prompt for Proprietary Models for BEnQA Zero-Shot Benchmark}
    \label{tab:zero-shot_prompt}
\end{table*}

\paragraph{Chain-of-Thought Prompting for BEnQA Few-shot benchmark: }
The prompt we use for few-shot benchmarking with the Proprietary models is shown in Table \ref{tab:few-shot_prompt}.

\begin{table*}[htbp]
    \centering
    \fbox{
        \parbox{\textwidth}{ 
            \textit{
            You are given a multiple choice question in English/Bengali.
            Your job is to answer it correctly. First reason step by step and then answer the question. \\ \\
            \{Question 1\}\\
            \{Answer 1\} \\ \\
            \{Question 2\} \\
            \{Answer 2\} \\ 
            - - - - - - - - - - - - - - - - - - - - - - - - - - - - - - - - - - - - - - - - - - - - - - - - - - - - - - - - - - - - - - - - - - - - - - - - - - - - - - - - - - - - - - - - - - - - - - - - - - - - - - - - - - - - - - - - - - - - - - - - - - - - - - - - - - - - - - - - - - - - - - - - - -\\
            \{Question\} \\
            }
        }
    }
    \caption{Prompt for Proprietary Models for BEnQA Few-Shot Benchmark}
    \label{tab:few-shot_prompt}
\end{table*}

\paragraph{Prompt for Translation Appended Benchmark: }
As described in \ref{appendix: extra_experiments}, we use the prompt shown in Table \ref{tab:append_prompt} to restrict the model to only reason in English in the translation-append experiment.

\begin{table*}[htbp] 
    \centering 
    \fbox{
        \parbox{\textwidth}{ 
            \textit{
    You are given a situation and its possible reason effect/answer in Bengali and your job is to correctly identify the reason effect/answer of the situation. \\ \\
    For your better understanding, English translation is also given. However, you must answer in Bengali only. \\
    Reason step by step in Bengali and only then give me the final answer. \\ \\
The question is given below:
            }
        }
    }
    \caption{Prompt for Zero-shot Append Experiment}
    \label{tab:append_prompt}
\end{table*}

\section{Benchmark Statistics}
\label{appendix: stats}
This section contains benchmark results by types and datasets used for our experiments
\label{sec:benchmark}

\paragraph{BEnQA Zero-shot Benchmark}
Table \ref{tab:BEnQA_Benchmark} shows the zero-shot benchmark results on BEnQA.

\paragraph{BEnQA Few-shot Benchmark without Chain-of-Thought Reasoning}
Table \ref{tab:BEnQA_Few-shots_no_cot} shows the few-shot benchmark results on BEnQA without Chain-of-Thought Reasoning.

\paragraph{BEnQA Few-shot Benchmark with Chain-of-Thought Reasoning}
Table \ref{tab:BEnQA_Few-shots_cot} shows the few-shot benchmark results on BEnQA with the use of Chain-of-Thought.

\paragraph{BIG-Bench-Hard Zero-shot Benchmark}
Table \ref{tab:BBH_benchmark} shows the zero-shot benchmark results on selected reasoning-based BIG-Bench-Hard datasets.

\begin{table*}[ht]
\centering

\begin{tabular}{llll|l}
\textbf{Subject} & \textbf{Factual Knowledge} & \textbf{Procedural \& Application} & \textbf{Reasoning} & \textbf{Total by Dataset} \\
\hline
12th Bio I & 288 & 12 & 15 & 315 \\
12th Bio II & 297 & 8 & 28 & 333 \\
12th Chem I & 229 & 85 & 58 & 372 \\
12th Chem II & 185 & 145 & 64 & 394 \\
12th Phy I & 115 & 161 & 32 & 308 \\
12th Phy II & 168 & 143 & 27 & 338 \\
12th Math I & 13 & 368 & 20 & 401 \\
12th Math II & 24 & 327 & 45 & 396 \\
10th Bio & 308 & 21 & 27 & 356 \\
10th Phy & 178 & 119 & 27 & 324 \\
10th Math I & 45 & 267 & 73 & 385 \\
10th Math II & 24 & 316 & 58 & 398 \\
10th Chem & 268 & 91 & 35 & 394 \\
8th Math & 30 & 139 & 45 & 214 \\
8th Sci & 194 & 26 & 13 & 233 \\
\hline
\textbf{Total} & \textbf{2366} & \textbf{2228} & \textbf{567} & \textbf{5161} \\
\hline
\end{tabular}
\caption{BEnQA Dataset Question Count by Subjects and Categories}
\label{category_Table}
\end{table*}

\begin{table*}[ht]
\centering
\small
\begin{supertabular}{llllllll}
\textbf{Language}                                & \textbf{Dataset}      & \textbf{GPT 4} & \textbf{GPT 3.5} & \textbf{Claude 2.1} & \textbf{LLaMA2 (13b)} & \textbf{LLaMA2 (7b)} & \textbf{Mistral 7b}  \\* 
\hline
\multirow{15}{*}{        English      } & 12th Bio I   & 84.44 & 69.21   & 59.37      & 38.10        & 30.48       & 31.43       \\*
                                        & 12th Bio II  & 81.98 & 63.96   & 55.26      & 33.63        & 24.62       & 38.14       \\*
                                        & 12th Chem I  & 82.57 & 57.37   & 52.06      & 24.13        & 19.35       & 27.08       \\*
                                        & 12th Chem II & 80.96 & 56.09   & 51.75      & 23.60        & 14.47       & 22.34       \\*
                                        & 12th Phy I   & 81.23 & 62.14   & 48.25      & 29.13        & 17.15       & 25.24       \\*
                                        & 12th Phy II  & 82.25 & 59.47   & 37.78      & 23.37        & 25.44       & 27.81       \\*
                                        & 12th Math I  & 85.54 & 59.10   & 57.93      & 10.22        & 6.23        & 11.22       \\*
                                        & 12th Math II & 77.02 & 53.28   & 56.51      & 16.92        & 6.82        & 15.15       \\*
                                        & 10th Bio     & 79.78 & 64.89   & 51.75      & 34.83        & 25.00       & 36.52       \\*
                                        & 10th Phy     & 78.70 & 62.46   & 53.02      & 28.92        & 19.38       & 26.77       \\*
                                        & 10th Math I  & 86.53 & 61.14   & 47.94      & 13.21        & 12.18       & 12.69       \\*
                                        & 10th Math II & 84.17 & 63.00   & 49.52      & 9.25         & 10.50       & 15.25       \\*
                                        & 10th Chem    & 86.55 & 62.44   & 56.83      & 28.68        & 22.08       & 28.43       \\*
                                        & 8th Math     & 85.05 & 70.09   & 56.07      & 26.17        & 21.03       & 21.50       \\*
                                        & 8th Sci      & 77.25 & 63.09   & 54.08      & 42.06        & 23.61       & 36.91       \\* 
\hline
\multirow{15}{*}{       Bengali       } & 12th Bio I   & 72.06 & 34.60   & 35.56      &              &             &             \\*
                                        & 12th Bio II  & 71.17 & 31.53   & 30.48      &              &             &             \\*
                                        & 12th Chem I  & 77.48 & 31.37   & 38.10      &              &             &             \\*
                                        & 12th Chem II & 75.13 & 35.53   & 34.60      &              &             &             \\*
                                        & 12th Phy I   & 77.35 & 37.54   & 42.54      &              &             &             \\*
                                        & 12th Phy II  & 77.51 & 31.95   & 29.84      &              &             &             \\*
                                        & 12th Math I  & 75.56 & 43.14   & 36.57      &              &             &             \\*
                                        & 12th Math II & 66.16 & 39.14   & 28.57      &              &             &             \\*
                                        & 10th Bio     & 77.53 & 35.11   & 41.85      &              &             &             \\*
                                        & 10th Phy     & 75.00 & 36.00   & 36.19      &              &             &             \\*
                                        & 10th Math I  & 77.46 & 40.67   & 34.92      &              &             &             \\*
                                        & 10th Math II & 78.14 & 38.25   & 35.87      &              &             &             \\*
                                        & 10th Chem    & 74.11 & 40.36   & 34.92      &              &             &             \\*
                                        & 8th Math     & 80.84 & 48.60   & 45.33      &              &             &             \\*
                                        & 8th Sci      & 72.10 & 35.62   & 35.62      &              &             &             \\
\hline

\end{supertabular}

\caption{BEnQA Zero-shot Benchmark}
\label{tab:BEnQA_Benchmark}
\end{table*}

\begin{table*}[ht]
\centering
\small
\begin{supertabular}{ll|cc|cc|cc}
\textbf{Language} & \textbf{Dataset} & \multicolumn{2}{c|}{\textbf{GPT 3.5}} & \multicolumn{2}{c|}{\textbf{LLaMA2 (7b)}} & \multicolumn{2}{c}{\textbf{LLaMA2 (13b)}} \\
& & \textbf{5-shot} & \textbf{3-shot} & \textbf{5-shot} & \textbf{3-shot} & \textbf{5-shot} & \textbf{3-shot} \\ 
\hline
\multirow{5}{*}{English}  & 10th Bio     & 58.26 & 59.13 & 48.70 & 43.48 & 43.48 & 39.13 \\ 
                          & 10th Phy     & 56.52 & 60.87 & 37.39 & 39.13 & 29.57 & 27.83 \\ 
                          & 10th Math I  & 41.74 & 43.48 & 30.43 & 25.22 & 33.91 & 28.70 \\ 
                          & 10th Math II & 46.09 & 37.39 & 21.74 & 21.74 & 40.00 & 42.61 \\ 
                          & 10th Chem    & 58.26 & 58.26 & 36.52 & 37.39 & 26.96 & 25.22 \\ 
\hline
\multirow{5}{*}{Bengali}  & 10th Bio     & 33.04 & 32.17 &      &      &      &      \\
                          & 10th Phy     & 41.74 & 35.65 &      &      &      &      \\
                          & 10th Math I  & 26.96 & 32.17 &      &      &      &      \\
                          & 10th Math II & 33.04 & 33.91 &      &      &      &      \\
                          & 10th Chem    & 41.74 & 41.74 &      &      &      &      \\
\hline
\end{supertabular}
\caption{BEnQA 10th Grade Few-shot Benchmark without Chain-of-Thought Reasoning}
\label{tab:BEnQA_Few-shots_no_cot}
\end{table*}

\begin{table*}[ht]
\centering
\small
\begin{supertabular}{ll|cc|cc|cc}
\textbf{Language} & \textbf{Dataset} & \multicolumn{2}{c|}{\textbf{GPT 3.5}} & \multicolumn{2}{c|}{\textbf{LLaMA2 (7b)}} & \multicolumn{2}{c}{\textbf{LLaMA2 (13b)}} \\
& & \textbf{5-shot} & \textbf{3-shot} & \textbf{5-shot} & \textbf{3-shot} & \textbf{5-shot} & \textbf{3-shot} \\ 
\hline
\multirow{5}{*}{English}  & 10th Bio     & 67.83 & 69.57 & 49.57 & 49.57 & 48.7 & 43.48 \\ 
                          & 10th Phy     & 66.09 & 70.43 & 40.00 & 38.26 & 38.26 & 43.48 \\ 
                          & 10th Math I  & 66.09 & 69.57 & 28.70 & 26.09 & 37.39 & 33.04 \\ 
                          & 10th Math II & 62.61 & 58.26 & 29.57 & 27.83 & 25.22 & 30.43 \\ 
                          & 10th Chem    & 68.70 & 63.48 & 44.35 & 42.61 & 43.48 & 42.61 \\ 
\hline
\multirow{5}{*}{Bengali}  & 10th Bio     & 36.52 & 33.91 &      &      &      &      \\
                          & 10th Phy     & 46.09 & 39.13 &      &      &      &      \\
                          & 10th Math I  & 34.84 & 41.74 &      &      &      &      \\
                          & 10th Math II & 40.87 & 45.22 &      &      &      &      \\
                          & 10th Chem    & 40.87 & 47.83 &      &      &      &      \\
\hline
\end{supertabular}
\caption{BEnQA 10th Grade Few-shot Benchmark with Chain-of-Thought Reasoning}
\label{tab:BEnQA_Few-shots_cot}
\end{table*}

\begin{table*}[h]
\centering
\small
\begin{supertabular}{lllll}
\textbf{Dataset}                & \textbf{English only} & \textbf{Bengali only} & \textbf{Gold Translation Append} & \textbf{GPT Translation Append} \\
\hline
Causal Judgement                & 57.2                 & 54.0                 & \textbf{58.8}                   & \uline{57.8}                    \\
Date Understanding              & 56.0        & 25.2                 & \textbf{43.2}                    & \uline{33.2}                            \\
Disambiguation QA               & 45.2                 & 40.0                 & \textbf{52.8}                   & \uline{51.2}                    \\
Formal Fallacies                & 55.2         & \textbf{53.6}        & \uline{47.6}                            & 46.0                            \\
Logical Deductions Five         & 35.2        & \textbf{21.6}         & 18.8                            & \uline{20.8}                            \\
Logical Deductions Seven        & 30.8        & 16.4                 & \textbf{22.8}                    & \uline{16.8}                            \\
Logical Deductions Three        & 51.6        & \uline{35.2}                 & 34.8                            & \textbf{36.0}                    \\
Multistep Arithmetic            & 68.5        & \textbf{49.0}         & \uline{43.5}                            &                                  \\
Navigate                        & 60.4        & \uline{55.6}                 & 55.2                            & \textbf{57.2}                    \\
Object Counting                 & 83.6        & 34.8                 & \uline{43.6}                            & \textbf{46.8}                    \\
Penguin In A Table              & 61.0        & 26.0                 & \textbf{36.3}                    & \textbf{36.3}                    \\
Reasoning About Colored Objects & 42.4        & 24.0                 & \textbf{32.4}                    & \uline{28.8}                            \\
Temporal Sequences              & 37.6        & \textbf{23.6}         & 22.8                            & \uline{21.2}                            \\
Web of Lies                     & 52.8        & 21.2                 & \textbf{48.8}                    & \uline{48.0}                            \\ 
\hline
\end{supertabular}
\caption{Big-Bench Hard Zero-shot Benchmark}
\label{tab:BBH_benchmark}
\end{table*}

\section{BEnQA Sample Questions}
\begin{figure*}[h]
    \centering
    \resizebox{0.95\textwidth}{!}{\includegraphics{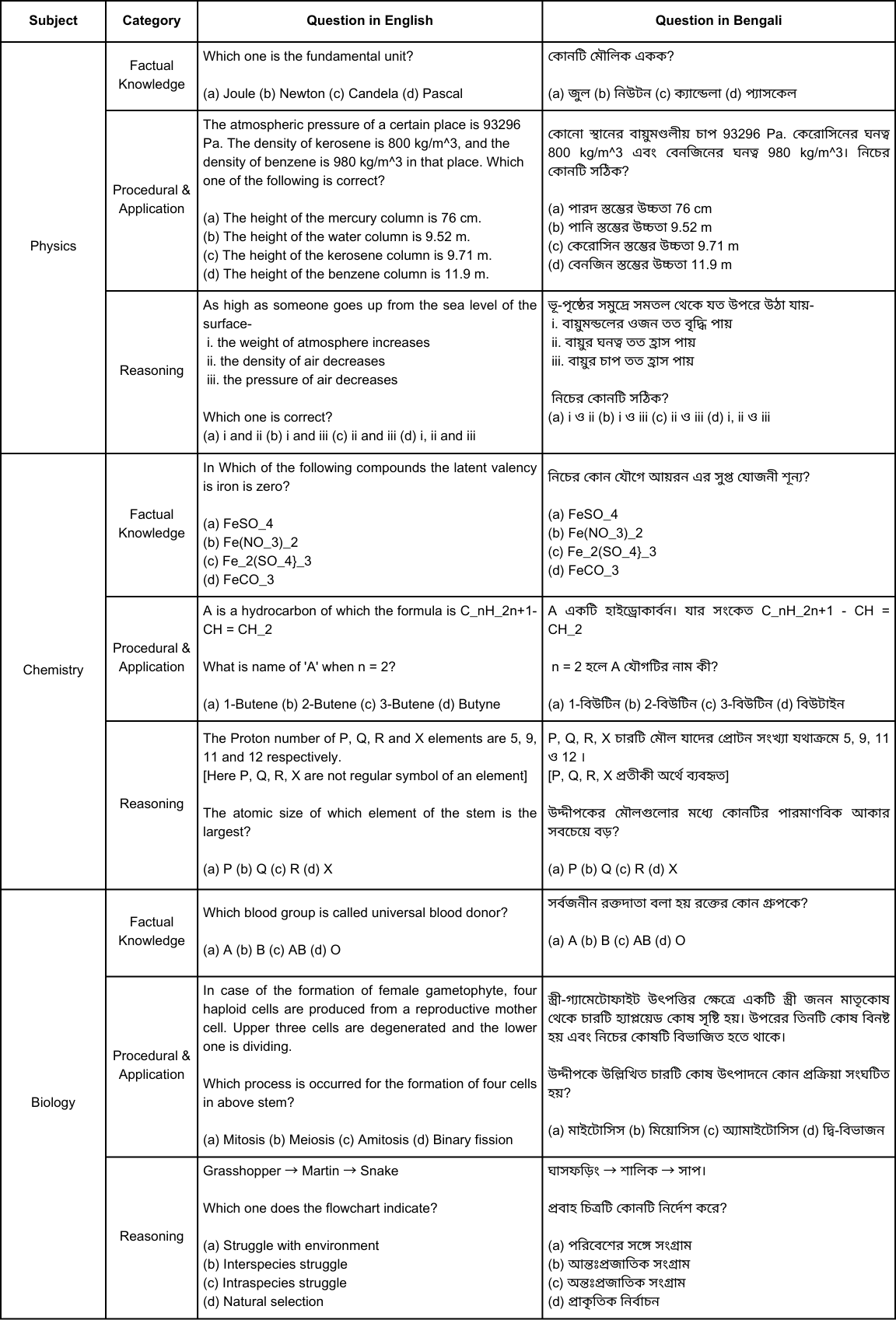}}
    \caption{BEnQA 10th Grade Sample Questions by Subject and Category}
    \label{fig:cat1}
\end{figure*}

\begin{figure*}[t]
    \centering
    \resizebox{0.95\textwidth}{!}{\includegraphics{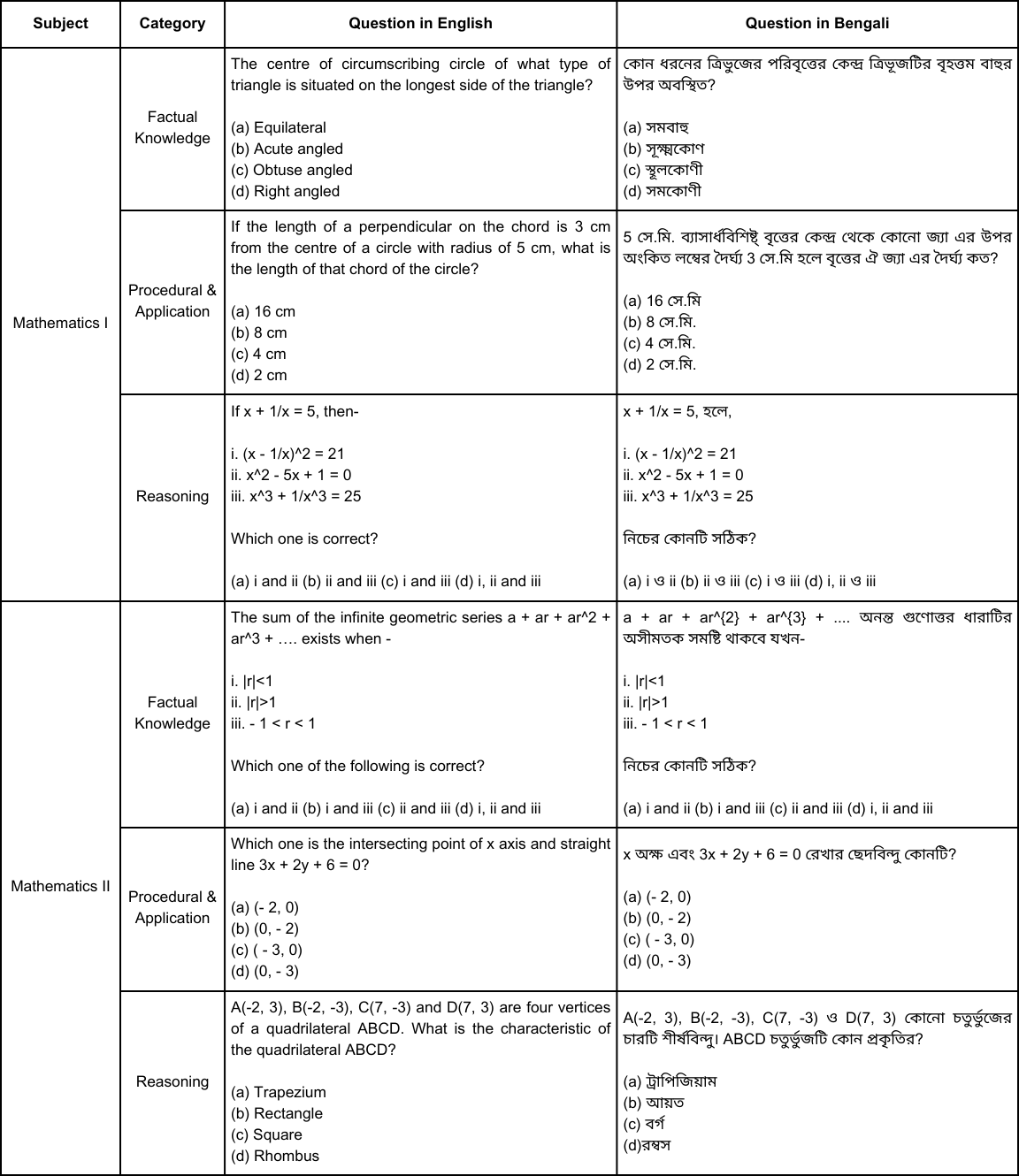}}
    \caption{BEnQA 10th Grade Sample Questions by Subject and Category}
    \label{fig:cat2}
\end{figure*}

We present some examples from the 10th grade subjects by the categories consisting both English and Bengali version of the in Figure \ref{fig:cat1} \& \ref{fig:cat2}. We kept the questions as it is, ie, we did not correct the subtle grammatical awkwardness in the English version of some of the questions.  

\end{document}